%% file: main.tex
\documentclass[manuscript, nonacm]{acmart}
\AtBeginDocument{%
  }

\usepackage{subfiles}
\usepackage{multicol, multirow, booktabs, bigstrut}
\usepackage{colortbl, nicematrix, longtable, hhline}
\usepackage[table]{xcolor}
\usepackage{enumitem}
\usepackage{booktabs}
\usepackage{todonotes}

\NewDocumentCommand{\matt}
{ mO{} }{\textcolor{teal}{\textsuperscript{\textit{Matt}}\textsf{\textbf{\small[#1]}}}}
\NewDocumentCommand{\sz}
{ mO{} }{\textcolor{blue}{\textsuperscript{\textit{Shuyan}}\textsf{\textbf{\small[#1]}}}}

\begin{document}

\title[Generalizability of Large Language Model-Based Agents: A Comprehensive Survey]{Generalizability of Large Language Model-Based Agents: \\ A Comprehensive Survey}

\author{Minxing Zhang}
\affiliation{%
  \institution{Duke University}
  \city{Durham}
  \state{North Carolina}
  \country{USA}
}
\email{minxing.zhang@duke.edu}

\author{Yi Yang}
\affiliation{%
  \institution{Duke University}
  \city{Durham}
  \state{North Carolina}
  \country{USA}}
\email{owen.yang@duke.edu}

\author{Roy Xie}
\affiliation{%
  \institution{Duke University}
  \city{Durham}
  \state{North Carolina}
  \country{USA}
}
\email{ruoyu.xie@duke.edu}

\author{Bhuwan Dhingra}
\affiliation{%
 \institution{Duke University}
 \city{Durham}
 \state{North Carolina}
 \country{USA}
 }
\email{bdhingra@cs.duke.edu}

\author{Shuyan Zhou}
\affiliation{%
  \institution{Duke University}
  \city{Durham}
  \state{North Carolina}
  \country{USA}}
\email{shuyan.zhou@duke.edu}

\author{Jian Pei}
\affiliation{%
  \institution{Duke University}
  \city{Durham}
  \state{North Carolina}
  \country{USA}}
\email{j.pei@duke.edu}

\renewcommand{\shortauthors}{Zhang et al.}

\begin{abstract}
Large Language Model (LLM)-based agents have emerged as a new paradigm that extends the capabilities of LLMs beyond static text generation to dynamic interaction with external environments. By integrating reasoning with perception, memory, and tool use, these agents are increasingly deployed in diverse domains such as web navigation, household robotics, finance, and healthcare. A critical challenge, however, lies in ensuring their \textit{generalizability} -- the ability to maintain consistently high performance across varied instructions, tasks, environments, and domains, especially those different from the agent's fine-tuning data. Despite growing interest, the concept of generalizability in LLM-based agents remains underdefined, and systematic approaches to measure and improve it are lacking. In this survey, we provide the first comprehensive review of generalizability in LLM-based agents. We begin by emphasizing the importance of agent generalizability by appealing to the agent's multiple stakeholders and clarifying the definition and boundaries of agent generalizability by situating it within a hierarchical domain-task ontology. We then review existing datasets, evaluation dimensions, and metrics, highlighting their strengths, limitations, and gaps. Next, we categorize strategies for improving generalizability into three groups: methods targeting the backbone LLM, approaches for agent specialized components, and techniques enhancing their interactions. Furthermore, we introduce the distinction between \textit{generalizable frameworks} and \textit{generalizable agents}, analyze their connections, and outline how generalizable frameworks can be translated into agent-level generalizability. Finally, we identify critical open challenges and future directions, including the development of standardized evaluation frameworks, variance- and cost-based metrics, and hybrid approaches that integrate methodological innovations with agent architecture-level designs. By synthesizing current progress and highlighting opportunities, this survey aims to establish a foundation for principled research on building LLM-based agents that generalize reliably across diverse real-world applications.
\end{abstract}



\maketitle

\input{sections_updated/introduction_updated}

\input{sections_updated/architecture_updated}

\input{sections_updated/measure_updated}

\input{sections_updated/improve_generalizability_llm}

\input{sections_updated/improve_generalizability_specialized_component}

\input{sections_updated/improve_generalizability_interaction}

\input{sections_updated/generalizable_method}

\input{sections_updated/conclusion_updated}

\begin{acks}
We thank Ruochen Huang from The Chinese University of Hong Kong, Shenzhen, who played a pivotal role in paper pool management. Through her diligent and meticulous approach, she curated a comprehensive collection of works on agent generalizability, systematically summarizing them across multiple dimensions including motivation, methodology, and evaluation metrics. She also critically analyzed the strengths and limitations of existing studies, offering comparative perspectives that were instrumental in identifying future directions. Her commitment and collaborative spirit made her an indispensable contributor, strengthening both the comprehensiveness and intellectual depth of this work.

Minxing Zhang, Yi Yang, and Jian Pei's research is supported in part in part by the NSF Project MSPA -2434666. All opinions, findings, conclusions and recommendations in this paper are those of the authors and do not necessarily reflect the views of the funding agencies. Jian Pei is an Amazon Scholar. Jian Pei holds concurrent appointments as a Professor of Computer Science at Duke University and as an Amazon Scholar. This paper describes work performed at Duke University and is not associated with Amazon. Roy Xie and Bhuwan Dhingra's research is supported in part by NSF award IIS-2211526 and by the Learning Engineering Virtual Institute under the grant number P0362134 SUB00004699. Roy Xie is also supported by the Apple Scholars in AIML Fellowship and the NSF Graduate Research Fellowship Program (GRFP).
\end{acks}

\bibliographystyle{ACM-Reference-Format}
\bibliography{sample-base}

\appendix

\end{document}

%% file: sections_updated/introduction_updated.tex
\section{Introduction}
Large Language Models (LLMs) have demonstrated impressive capabilities across a broad range of tasks, such as  natural language understanding~\cite{zhu2023minigpt}, code generation~\cite{nam2024using}, math reasoning~\cite{ahn2024large}, and creative writing~\cite{gomez2023confederacy}, etc. However, their static knowledge base and lack of real-time interactivity limit their utility in tasks that require dynamic engagement with external environments. To overcome these constraints, LLM-based agents have emerged as a new paradigm -- systems that combine the reasoning abilities of LLMs with the capacity to interact with and alter their environment.

An \textbf{LLM-based agent}, a term used interchangeably with ``agent'' throughout this survey, is an AI system centered around an LLM, augmented with additional components that enable it to perceive, act upon, and respond to external environments. Unlike standard LLMs, which provide purely textual outputs, these agents can perform real-world tasks by executing actions that modify the environment and incorporating feedback into subsequent decisions. For instance, while an LLM may explain how to book a flight, an LLM-based agent can autonomously navigate airline websites, complete forms, and finalize transactions -- activities requiring direct interaction with external systems.

\begin{figure*}[t]
    \centering
    \includegraphics[width=\textwidth]{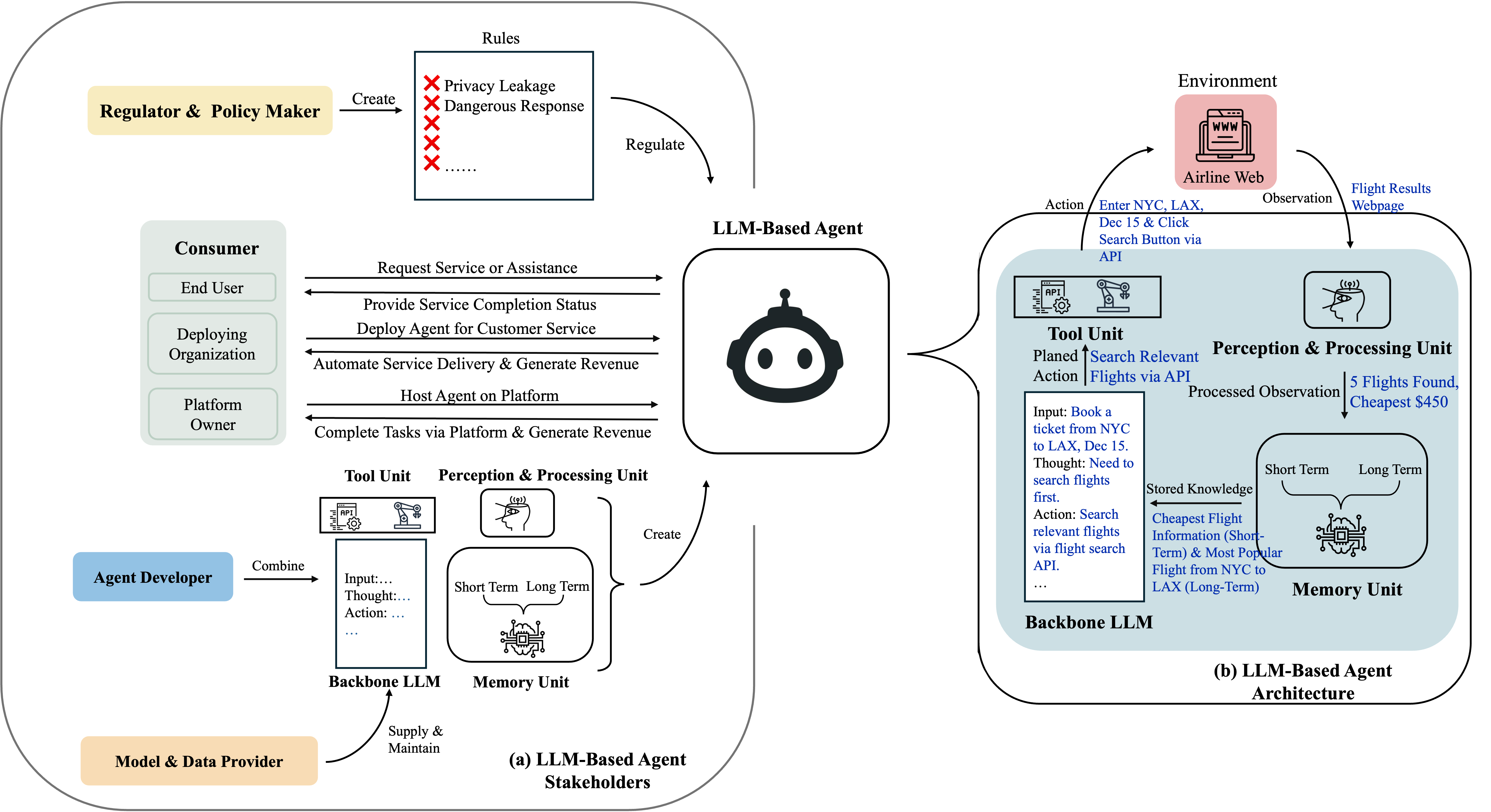}
    \caption{LLM-based agent ecosystem and architecture. (a) Shows different stakeholders and their interactions with the agent, including regulators and policy makers, consumers (end users, deploying organizations, platform owners), agent developers, and model and data providers. (b) Shows the agent architecture using an airline ticket booking example to concretely illustrate the workflow (concrete examples shown in dark blue as illustrations).}
    \label{fig:llm_agent_figure}
\end{figure*}

To carry out such actions, an LLM-based agent must coordinate several core components. Consider an LLM-based agent designed for air ticket booking, illustrated in Figure~\ref{fig:llm_agent_figure}(b): an LLM receives user instructions and serves as the reasoning engine, planning a sequence of actions such as searching for flights, applying user preferences, and completing purchases. A tool unit executes these plans by interacting with external systems (e.g., airline websites) through API calls; when navigating airline websites, a perception and processing unit observes and interprets complex textual or visual data such as seat maps and pricing information; and a memory unit stores observations and the agent's associated thoughts and actions for context-aware decision-making for the future, such as recalling recent flight search queries (short-term memory) and the most popular flights rated by users (long-term memory). The agent's performance depends on the integration of these components -- reasoning, perception, action, and memory -- toward a unified goal.

Thanks to these capabilities, LLM-based agents extend traditional LLMs and are increasingly deployed across diverse domains. Applications include web-based tasks such as ticket booking, profile searching, and multi-site web browsing~\cite{deng2024mind2web}, as well as embodied tasks like object manipulation in household environments~\cite{shridhar2020alfworld}. As their use broadens, generalizability becomes a critical concern. 
In this context, \textbf{generalizability refers to an agent's ability to apply its core functionalities, like planning, tool calling, knowledge retrieval, and observation processing, to maintain consistently high performance across different user instructions, environments, tasks, or domains}~\cite{stahl2020expanding, yang2024agentoccam}. Two important aspects characterize this definition. First, generalizability requires consistently high performance across different dimensions (i.e., user instructions, tasks, environments, or domains); for instance, a generalizable agent should demonstrate consistently high performance like accuracy regarding ticket booking across different airline websites, from premium carriers like Delta and American Airlines to budget options like Frontier, despite their varying interface designs and booking workflows. Second and moreover, generalizability emphasizes consistently high performance on scenarios absent from the agent's fine-tuning data. That is, since agents are typically fine-tuned on specific downstream tasks using specific datasets, generalizability means extending beyond this fine-tuning coverage. For example, a generalizable agent fine-tuned only on Delta's interface should successfully navigate Frontier's, handling differences in pricing structures, seat selection processes, and payment procedures without prior exposure to these variations.
In short, generalizability reflects an agent's capability to reliably complete user-specified tasks across diverse, unseen scenarios~\cite{bruton2000reliability}.

Generalizability is closely tied to user trust: stakeholders are more likely to rely on agents that consistently perform as expected across varying conditions (further discussed in Section~\ref{sec:agent_stakeholder}). Its importance is magnified in high-stakes domains such as finance, healthcare, and autonomous systems. For example, without generalizability a financial agent trained on U.S. market data may incorrectly apply domestic trading strategies to international markets, leading to regulatory violations or financial loss. Similarly, a diagnostic agent trained on a narrow demographic (e.g., young Caucasian males) may produce inaccurate diagnoses when applied to different populations (e.g., elderly Asian females), due to variations in symptom presentation. These risks highlight the need to identify and address the unique challenges of generalizability in LLM-based agents -- challenges that go beyond those faced by static LLMs.

\subsection{Challenges and Contributions}

Despite growing recognition of its importance, achieving generalizability in LLM-based agents remains challenging.

\subsubsection{Challenges} 
\label{sec:intro_challenges}

First, \emph{a key obstacle is the lack of a standardized, universally accepted framework to define the boundaries of agent generalizability.} Such a framework could adopt a hierarchical structure, resembling a domain-task ontology where broad domains are decomposed into sub-domains and increasingly specialized task categories. Within this structure, an agent's generalizability could be explicitly defined as the region in which it consistently demonstrates high performance, thereby delineating the scope of its generalizable capabilities. 
\emph{To construct such a standardized domain-task ontology, researchers could leverage established industry classification systems rather than ad-hoc categorizations.} For instance, the North American Industry Classification System (NAICS)~\cite{naics2022} provides a rigorous hierarchical framework that categorizes economic activities into sectors such as Information, Education, and Healthcare. Within the Information sector, there are subsectors like Publishing Industries and Telecommunications, which further decompose into industry groups such as Computing Infrastructures, and ultimately into specific industries like Web Hosting. Within each NAICS-derived industry, existing agent benchmarks can provide task-level classifications -- Mind2Web~\cite{deng2024mind2web} demonstrates this approach by organizing web-based tasks hierarchically by user intents (e.g., booking, purchasing), objects (e.g., flights, products), and conditions (e.g., dates, preferences), ensuring that agent generalizability claims are grounded in established industrial standards and empirically validated task hierarchies rather than subjective categorizations. Such standardization would enable fair comparisons across agents by standardizing the levels at which generalizability is evaluated. In the absence of such a framework, studies often make inconsistent claims about generalizability, each using different levels of granularity. For example, one agent may claim generalizability by performing object manipulation tasks (e.g., pick-up, place, open, close) across various indoor environments~\cite{zhang2024lamma}, another by executing similar tasks in both indoor and outdoor settings~\cite{sun2024trustnavgpt}, and a third by performing heterogeneous web-based tasks across sub-domains like travel, information services, and shopping~\cite{deng2024mind2web}. While each claims ``generalizability,'' their generalizability boundaries and evaluation dimensions differ significantly. Once such a standardized framework is established, studies can no longer vaguely declare that their agents are generalizable without precisely positioning their agent within the structured domain-task hierarchy -- for example, ``web-based domain $\rightarrow$ travel $\rightarrow$ ticket booking'' -- thereby eliminating ambiguity and enabling fair comparison.

Second, \emph{another core challenge is establishing quantitative measurement, theoretical guarantees, and practical methods to improve agent generalizability.} Measurement requires comprehensive, unbiased benchmarks, including evaluation metrics, datasets, and task sets. For example, in airline ticket booking, generalizability might be measured by how consistently the agent selects optimal actions -- such as identifying the lowest-cost flights -- across user instructions from diverse demographic groups, benchmarked against expert annotations. With such evaluations in place, generalizability guarantees could be formulated, e.g., ensuring the agent successfully completes booking tasks on at least $80\%$ of airline websites without human intervention. These guarantees can be further improved through methods like increasing training data diversity~\cite{zeng2023agenttuning}, such as incorporating data from multiple airline platforms. Enhancing generalizability may involve improvements to each agent component -- the LLM, perception unit, memory, and tool interfaces -- as well as better coordination among them. A taxonomy of such methods would support more systematic advancement.

Third, \emph{there remains a conceptual and practical gap between generalizable frameworks and generalizable agents.}
In this survey, we define a generalizable framework as one that yields consistently high performance regardless of instruction, environment, task, or domain type, when combined with associated agent's fine-tuning data. For instance, a framework that, when combined with household fine-tuning data, enables an agent to achieve strong test performance on household tasks, similarly enables the agent to achieve strong results in web-based tasks when combined with web-based fine-tuning data. In other words, a generalizable framework, when combined with a fine-tuning dataset for any specific scenario (i.e. user instruction, task, environment, or domain), can enable the agent to achieve high performance like accuracy in that scenario. Examples include frameworks such as ReAct~\cite{yao2022react} and Reflexion~\cite{shinn2024reflexion}. However, deploying a generalizable framework does not guarantee the resulting agent will itself be generalizable. We define a generalizable agent, as earlier, as one that maintains consistently high performance across varied instructions, tasks, environments, or domains, especially those not seen during agent fine-tuning. For example, an agent fine-tuned on household object pickup tasks should generalize to related tasks such as object placement, opening, and closing, or even to other sub-domains like cooking or cleaning without being fine-tuned on those tasks. The key distinction is that: a generalizable framework with fine-tuning data for any scenario can enable the agent to achieve high accuracy on that specific scenario, but may not on other scenario different from the fine-tuning dataset. In contrast, a generalizable agent can achieve consistent accuracy across different scenarios, especially those that differ from its fine-tuning dataset. With this, the core challenge, then, is understanding how to translate generalizable frameworks into generalizable agents -- i.e., how methodological generalizability can be leveraged to ensureconsistent agent behavior across diverse scenarios.

\subsubsection{Our Contributions} 

This survey provides a comprehensive review of the literature on the generalizability of LLM-based agents, synthesizes the current state of the field, and addresses the three fundamental challenges.

First, we identify a key gap in the formal definition of generalizability and the absence of a standardized framework for its evaluation. To address this, we propose a formal definition: LLM-based agent generalizability is the agent's ability to maintain consistently high performance across varied user instructions, environments, tasks, and domains, especially those that differ from its fine-tuning data.
We emphasize the importance of a universally accepted framework to define the scope and boundaries of generalizability. A hierarchical taxonomy -- spanning broad domains (e.g., web-based, household) to sub-domains (e.g., travel, shopping) to fine-grained tasks (e.g., air ticket booking, appointment scheduling) -- can serve as the foundation for standardizing generalizability and enabling meaningful comparisons across agents.

Second, although prior studies have explored aspects of the second challenge -- such as proposing evaluation metrics, constructing benchmarks, or designing methods to enhance generalizability -- there remains no unified, systematic survey that integrates and analyzes these contributions. We address this by introducing a structured taxonomy that covers both measurement and improvement. We review existing evaluation metrics, datasets, and task setups for assessing generalizability, analyzing their strengths and limitations. We then categorize improvement strategies into three areas: methods that target the backbone LLM, those that focus on agent specialized components (e.g., perception, memory, tools), and those that enhance the interaction between the LLM and agent specialized components. For each, we provide an analysis of current methods and highlight key insights and open research questions to guide future work.

Third, we examine the often-overlooked distinction between generalizable frameworks and generalizable agents. We formally define both and clarify their differences. Generalizable frameworks -- such as ReAct~\cite{yao2022react} and Reflexion~\cite{shinn2024reflexion} -- are typically designed to be robust across scenarios when combined with scenario-specific fine-tuning data, yet their application to developing agents with robust generalizability remains underexplored. We analyze how such frameworks can contribute to building agents that generalize across different scenarios, and emphasize the need for principled strategies that link methodological innovation to agent-level generalizability.

This work makes several novel contributions. To the best of our knowledge:
\begin{itemize}
    \item It provides the first formal definition of LLM-based agent generalizability and proposes a structured taxonomy for its measurement and improvement;
    \item It is the first to examine agent generalizability from the perspective of multiple stakeholders, motivating why it matters across different usage contexts;
    \item It introduces the first comprehensive analysis of agent generalizability at the component level, highlighting how individual agent modules and their interactions affect generalizability;
    \item It is the first to define and compare generalizable frameworks and generalizable agents, and to explore the relationship between them; and 
	\item It synthesizes current research trends, identifies strengths and limitations in existing work, and outlines critical open problems and future directions for advancing the field.
\end{itemize}

\subsubsection{Article Organization} 
The remainder of this paper is organized as follows. We begin by introducing the architecture and ecosystem of LLM-based agents in Section~\ref{sec:SA_architecture}, detailing the functionality of each component and the roles of various stakeholders, and explaining why generalizability is essential from each stakeholder's perspective. We then review existing approaches to measuring agent generalizability in Section~\ref{sec:measure_generalizability}, including datasets, tasks, and evaluation metrics, followed by a structured discussion of their limitations and promising future directions. Then, we discuss approaches for improving agent generalizability: these approaches are categorized based on whether they target the backbone LLM itself (Section~\ref{sec:improve_generalizability_LLM}), the agent specialized components (Section~\ref{sec:improve_generalizability_agent_components}), or their interactions (Section~\ref{sec:improve_generalizability_interaction}). Each group of methods is examined to identify strengths, weaknesses, and open research directions. We then turn to the distinction between generalizable frameworks and agents in Section~\ref{sec:generalizability_method_vs_generalizable_agent}, summarizing current generalizable frameworks, categorizing them by approach, and analyzing how they can be leveraged to construct more generalizable agents. In key areas such as instruction optimization and intent modeling, we also assess current approaches and highlight underexplored research opportunities. Finally, we conclude with a summary of key insights and future research directions in Section~\ref{sec:conclusion_updated}.

\subsection{Distinctions from Related Surveys}

While several recent surveys have addressed aspects of LLM-based agents~\cite{agentsurvey2025, huang2024understanding, zou2025surveylargelanguagemodel, masterman2024landscape, jiang2024multi, tang2024prioritizing, xi2023rise}, they fall short in several key areas. This survey offers a more comprehensive and structured treatment of generalizability in LLM-based agents, addressing notable gaps across three dimensions.

\textbf{Comprehensive Exploration of Agent Generalizability with Formal Definition.} Existing surveys have broadly examined agent performance in terms of accuracy~\cite{bratko1997machine, huang2024understanding, masterman2024landscape}, security~\cite{tang2024prioritizing, xi2023rise}, privacy~\cite{li2024personal, he2024emerged}, and fairness~\cite{xie2024can, borah2024towards}. However, generalizability has received limited attention. In particular, there are a) no formal definition of agent generalizability, b) no established benchmarks for fair, standardized comparisons, and c) no comprehensive summary of methods aimed at improving it. These gaps lead to inconsistent claims in the literature, where generalizability is asserted across vastly different task boundaries and granularities. Additionally, most existing surveys organize findings by methodology -- e.g., categorizing by planning, memory, or action mechanisms~\cite{agentsurvey2025, huang2024understanding} -- without specifying the performance objective each method is intended to improve. That is, they do not answer the fundamental question of ``which specific aspects of performance (whether accuracy, security, privacy, or others) each category of methods is designed to improve.'' This lack of alignment between methods and performance goals creates ambiguity. In contrast, our survey focuses specifically on generalizability, introduces the first formal definition for it in the context of LLM-based agents, and proposes a structured taxonomy of improvement strategies explicitly aimed at enhancing this dimension.
	
\textbf{Component-Level Analysis of Agent Generalizability.} Existing surveys often treat LLM-based agents as monolithic systems or focus exclusively on the backbone LLM, overlooking the role of agent specialized components such as perception, memory, and tool modules and their interactions. For instance, \citet{tang2024prioritizing} and \citet{xi2023rise} primarily focus on limitations and mitigation strategies for general-purpose LLMs (e.g., hallucinations, outdated knowledge), without sufficiently addressing the unique challenges posed by other agent components or their interactions. While these works occasionally mention perception, tools, or memory, they fall short of analyzing how to systematically improve generalizability across these components or how their interactions influence agent behavior. In contrast, our survey explicitly examines generalizability at the component level, emphasizing that agent performance depends on more than just the LLM. We analyze how generalizability challenges emerge from different modules and their coordination, offering a more holistic view of agent architecture.
	
\textbf{Formal Definition and Analysis of Generalizable Frameworks versus Generalizable Agents.} Existing surveys categorize frameworks based on functionality but fail to examine whether the frameworks themselves are generalizable or, more specifically, which dimension (i.e., instruction, task, environment, or domain) each framework can generalize to. Some frameworks, such as ReAct~\cite{yao2022react} and Reflexion~\cite{shinn2024reflexion}, have demonstrated strong generalizability across different environments, yet the literature lacks a formal framework for defining or identifying such generalizable frameworks. Without understanding framework-level generalizability (i.e., the specific dimension the framework can generalize to), practitioners cannot effectively select and adapt these proposed frameworks for their specific use cases. Furthermore, current work does not distinguish between generalizable frameworks and generalizable agents, nor does it explore how the former can be systematically leveraged to construct the latter. Our survey addresses this gap by formally defining both terms, clarifying their distinctions, and proposing how generalizable frameworks can be leveraged to create more generalizable agents.

%% file: sections_updated/architecture_updated.tex
\section{LLM-based Agents and Ecosystems}
\label{sec:SA_architecture}

This section introduces the architectural foundations and ecosystem context of LLM-based agents. We begin by presenting a modular view of agent architectures, describing the backbone LLM and its specialized components for perception, memory, and tool use. We then discuss current architectural limitations that hinder generalizability, identifying two primary challenges: insufficient communication between components and inadequate orchestration across heterogeneous modules. Finally, we examine multi-stakeholder perspectives -- covering consumers, developers, providers, and regulators -- to highlight how different ecosystem participants evaluate and influence agent generalizability.

\subsection{LLM-based Agent Architecture Overview}
\label{sec:agent_architecture_subsection}

Similar to~\citet{russell1997rationality} and~\citet{weng2023agent}, we define an LLM-based agent as a system composed of an LLM (referred to as \textbf{the backbone LLM} in later discussions to emphasize that it is specifically used for agents) -- responsible for language understanding and decision-making -- and a set of specialized components that enable the agent to interact with, modify, and perceive its external environment. 

These specialized components -- the perception and processing unit, memory unit, and tool unit -- are purpose-built modules that extend the backbone LLM's capabilities to act and learn in dynamic environments. Thus, we refer to these collectively as \textbf{agent specialized components}. For brevity, we will use the term ``agent components'' and ``agent specialized components'' interchangeably throughout this survey. A high-level illustration is shown in Figure~\ref{fig:llm_agent_figure}.

Before detailing the functions of each component, let us briefly introduce two representative environments that LLM-based agents operate in, both of which demonstrate how agents perceive observations and perform actions that alter environmental states.

\begin{example}[Representative LLM-based agent environments]\label{ex:environments}
First, let us consider web search tasks. In web-based environments~\cite{deng2024mind2web, yao2022webshop, yang2024agentoccam}, agents interact with websites and documents to fulfill user queries. They modify the environment by submitting search queries, clicking links, and navigating pages, while receiving feedback through returned search results, webpage content changes, error messages for broken links, or updated browser history.

As another example, consider household tasks. In physical settings such as smart homes~\cite{sun2024trustnavgpt, shridhar2020alfworld, zhang2024lamma}, agents interpret natural language commands and perform actions like object manipulation or appliance control. They affect the environment by moving items, adjusting device settings, or changing physical conditions like temperature or humidity, and receive feedback via sensors, visual confirmation of task completion, or user responses.
\qed
\end{example}

The examples in Example~\ref{ex:environments} reflect a common interaction loop: user instructions are processed by the agent, which generates plans and executes actions; the environment responds with updated observations, which are then perceived and processed by the agent to guide future decisions.

\subsubsection{Components of LLM-Based Agents}

The agent architecture consists of the backbone LLM and four core agent components, each playing a distinct functional role:

\paragraph{Backbone LLM} The backbone LLM serves as the reasoning and planning core~\cite{xi2025rise}. It interprets user instructions, generates high-level plans, coordinates interactions among agent components, and produces natural language outputs or API-like function calls for downstream execution.

\paragraph{Perception and Processing Unit} This unit captures and processes observations from the environment. In physical domains (e.g., autonomous driving), it may include perception modules like object detectors using LiDAR~\cite{song2024enhancing}. Additionally, the it processes observations to convert them into structured format as input to the backbone LLM. For instance, in web-based tasks, it transforms lists blocks or tables into Markdown~\cite{yang2024agentoccam}. Typically, it may apply compression or summarization techniques to ensure token efficiency, e.g., by removing redundant HTML elements from web pages.

\paragraph{Memory Unit} The memory unit provides both short-term and long-term memory functionalities. Short-term memory includes the agent's previous thoughts, actions, and associated observations within the current session, while long-term memory may consist of external knowledge bases where task-relevant documents are retrieved based on user instructions. This memory allows the agent to reference past experiences and external information, thereby supporting planning continuity and informed decision-making~\cite{sumers2023cognitive, kotseruba202040}.

\paragraph{Tool Unit} The tool unit enables the agent to combine the backbone LLM's planning capabilities with concrete functional abilities to achieve its objectives. According to~\citet{wang2024tools}, tools are function interfaces (e.g., APIs, software commands, or physical actuators) to computer programs that run externally to the backbone LLM, where the backbone LLM generates function calls and input arguments to use them. From the perspective of the backbone LLM, the actions it plans are essentially API calls, each consisting of a specific function name and the corresponding arguments, even though the actual API implementations encode different levels of operational granularity. For instance, both invoking a calculator (\texttt{calculator(formula=``2+2'')}) and executing a movement action (\texttt{move(direction=``forward'')}) can be represented as API calls, despite their differences in granularity. These APIs abstract the underlying implementations, translating the calls into operations that produce concrete effects on the environment.

Together, these components form a feedback-driven, modular system in which the backbone LLM guides planning and reasoning, while agent specialized components enable perception, memory integration, and action execution. This modular design is central to enabling generalizable and adaptive behavior across tasks and environments.

\subsubsection{Limitations in LLM-Based Agent Architectures and Future Directions}
Current LLM-based agent architectures face several practical limitations that constrain both generalizability and broader aspects of agent performance, including accuracy, security, privacy, and fairness. We categorize these architectural challenges into two main areas: a) insufficient communication between agent components and b) inadequate orchestration of heterogeneous components. Addressing these issues presents promising directions for improving agent generalizability.

\paragraph{Insufficient Communication Between Agent Components}  
One key limitation lies in the lack of effective communication channels between agent components, which can lead to inconsistent behaviors or degraded performance. For example, \citet{yang2024agentoccam} and \citet{ji2024dynamic} emphasize that the perception and processing unit should summarize or compress observations to promote generalizability and reduce context length, thus improving inference efficiency. However, as noted by \citet{wang2024safe}, such compression may result in the loss of critical details, potentially causing the backbone LLM to generate unsafe or inaccurate actions due to incomplete understanding of the environment.

These conflicting objectives are both valid: the perception and processing unit aims to optimize generalizability and efficiency, while the backbone LLM requires detailed observations to ensure high task-specific accuracy and adherence to security protocols. The absence of explicit communication between components about their intents and internal strategies (e.g., summarization heuristics or compression levels) prevents coordinated adaptation, leading to suboptimal agent behavior. Similar issues may arise in transferring knowledge from the memory unit to the backbone LLM, where incompatible representations or objectives can hinder effective retrieval and use~\cite{hu2024hiagent, yang2024agentoccam}.

A promising direction is to enhance intra-agent communication protocols that allow components to share operational intents or assumptions, enabling them to adjust their operations accordingly. For instance, if the perception and processing unit applies generalized representations (e.g., abstract summaries of visual content), it should signal this to the backbone LLM, which can then adapt its planning strategies to account for potential information loss. Such communication would support flexible prioritization (e.g., generalizability vs. accuracy) depending on task requirements, ultimately improving coordination and agent performance.

\paragraph{Inadequate Orchestration of Heterogeneous Components}  
Another limitation is the assumption that the backbone LLM can effectively coordinate all agent components -- even when those components originate from different providers and follow incompatible specifications. This challenge of orchestrating heterogeneous modules is particularly evident in real-world deployments, where agent developers assemble agents using third-party LLMs, tools, memory systems, and perception modules~\cite{agentUmf, anthropic2024mcp, langchain2022}.

\begin{example}\label{ex:heterogeneiousComponents} 
An LLM from Provider $A$ may not inherently understand the organizational structure of a tool unit from Provider $B$ (e.g., web browsing functions are organized by interaction type -- scrolling, clicking, typing -- rather than by website categories), the compressed representations used by a perception module from Provider $C$ (e.g., representing visual scenes as abstract semantic graphs rather than natural language descriptions), or the query parameters used by a memory system from Provider $D$ (e.g., nearest neighbor thresholds for retrieval)~\cite{asai2023self, wang2024safe, yang2024agentoccam}.

Existing mitigation strategies -- such as directly prompting the backbone LLM about agent components' functionality like tool descriptions~\cite{shi2025prompt}, or fine-tuning the LLM on component-specific datasets like training the backbone LLM with (query-knowledge) pairs using the specific querying language supported by the memory unit's DBMS -- have practical limitations. Prompting can increase inference latency and is sensitive to prompt structure~\cite{chuang2024learning, shi2025tool}; meanwhile, fine-tuning may not be feasible due to model accessibility constraints, lack of component-specific training data, or the risk of catastrophic forgetting~\cite{wang2024comprehensive}.
\qed
\end{example}

Example~\ref{ex:heterogeneiousComponents} highlights the importance of communication and coordination of heterogeneous agent components in an effective and efficient manner. While standardized communication protocols like MCP~\cite{anthropic2024mcp} offer partial solutions for basic information exchange between components, challenges still remain. That is, besides basic information exchange, there is still need for better coordination among heterogeneous agent components to support adaptive information routing (e.g., deciding which observations should be passed to the backbone LLM and when), automated tuning of component parameters (e.g., memory retrieval thresholds) based on specific contexts, and compatibility problems when new components replace old components (for instance, the tool calling template adopted by ToolLLM~\cite{qin2023toolllm} differs from that of ToolFormer~\cite{schick2023toolformer}, which complicates reusing ToolLLM with resources designed for ToolFormer)~\cite{shi2025tool}.

Given this, a promising direction emerges: the development of a \textit{coordination mechanism} that explicitly manages interactions between the backbone LLM and agent specialized components. Such a mechanism could mediate configuration mismatches, handle representation translation, and optimize information flow without requiring further customization of the backbone LLM. 
For instance, this mechanism can be an explicit central coordination unit (assuming this coordination has full information of capabilities provided by different component providers, acting like a trusted third party) that takes heterogeneous components as input and outputs the way to solve the incompatibility, such as an optimal organization of tools in the tool unit, optimal summarization techniques, or the optimal communication among components (e.g., how the planning instructions are optimally delivered from the backbone LLM to the tool unit, or in what formats summarized observations are transferred back to the backbone LLM). While recent work (e.g., \citet{agentUmf}) has introduced coordination units for managing LLM-environment interaction,
their functionality largely overlaps with existing components like the perception and tool units. What remains unexplored is how to design a coordination mechanism that addresses \textit{component heterogeneity}, not just environmental interfacing.

With this, determining how to \textit{train} this coordination unit remains an open question -- should it be optimized end-to-end with the rest of the agent? Should it use supervised labels, reinforcement signals, or meta-learning strategies? An intuitive way to train this unit is to use a large number of heterogeneous components, such as different backbone LLMs, memory systems, and perception mechanisms, with the optimal configuration as ground truth. 

Addressing these architectural limitations by improving inter-component communication and designing dedicated coordination mechanisms holds the potential to substantially enhance the generalizability and more broadly, performance, of LLM-based agents across complex, heterogeneous environments.

\subsection{Multi-Stakeholder Perspectives on LLM-Based Agent Generalizability}
\label{sec:agent_stakeholder}

LLM-based agent generalizability is valued by a wide range of stakeholders, each with distinct motivations and operational contexts. We group these stakeholders into four categories: consumers, model \& data providers, agent developers, and regulators \& policy-makers. This section explores their perspectives and the implications of agent generalizability for each group. Table~\ref{tab:stakeholder_generalizability} summarizes the discussion.

\begin{table*}[t!]
\centering
\caption{Stakeholder Perspectives on LLM-Based Agent Generalizability}
\label{tab:stakeholder_generalizability}
\begin{tabular}{p{1.8cm}p{3.8cm}p{3.3cm}p{5cm}}
\toprule
\textbf{Stakeholder} & \textbf{Role} & \textbf{Motivations} & \textbf{Why Generalizability Matters} \\
\midrule

\textbf{End Users} & Individuals or organizations directly using agents for personal or operational tasks & Convenience, cost-efficiency, access to multi-functional services & Reduces need for multiple agents, streamlines interaction, improves experience across varied user tasks \\
\midrule

\textbf{Deploying Organizations} & Companies or institutions integrating agents into products or workflows & Revenue generation, cost reduction, scalability & Allows reuse of the same agent across strategic shifts and diverse service domains, minimizing retraining and deployment costs \\
\midrule

\textbf{Platform Owners} & Platforms and environments hosting agent operations (e.g., websites, smart systems) & Ecosystem control, platform compatibility, user satisfaction & Ensures agents remain functional despite interface or infrastructure changes, reducing maintenance overhead and developer friction \\
\midrule

\textbf{Model Providers} & Developers of backbone LLMs used in agent systems & Model adoption, ecosystem growth, standardization & Agent-friendly models are more integrable and resilient, increasing adoption and reducing customization costs \\
\midrule

\textbf{Data Providers} & Curators and suppliers of training datasets for LLMs & Dataset usage, benchmarking leadership, market demand & Generalizable agents increase demand for well-curated, domain-spanning datasets; promote long-term dataset relevance \\
\midrule

\textbf{Agent Developers} & Engineers assembling full agents from LLMs and agent specialized components & Performance optimization, cost efficiency, market expansion & Enables scaling across tasks and clients, avoids repeated post-training, supports modular and reusable system design \\
\midrule

\textbf{Regulators, }\textbf{Policy-makers} & Authorities enforcing compliance, security, and fairness in agent deployment & Public trust, legal enforcement, ethical AI deployment & Generalizability ensures consistent agent performance across demographics and domains, supports fairness and privacy\\
\bottomrule

\end{tabular}
\end{table*}

\subsubsection{Consumers}

Consumers include individuals or organizations that use LLM-based agents to receive services and accomplish specific objectives, such as travel booking, diagnostics, or workflow automation. Agents are generally viewed as black boxes, with expectations focused on output quality rather than internal mechanisms. Consumers are further categorized into end users, deploying organizations, and platform owners.

\paragraph{End Users} 
These are individuals or institutions who directly engage with agents to fulfill their own needs. For instance, individuals provide inputs (like destination, passport number, and travel dates) to LLM-based agents for ticket booking and receive outputs like confirmed reservations, itineraries, or payment receipts. Generalizability is critical for satisfy end users' diverse requests across different domains, addressing a fundamental efficiency concern -- users prefer interacting with a single agent rather than managing multiple specialized agents for different tasks~\cite{clarke2024one}. For example, students prefer agents that can solve math problems, guide science experiments, and offer writing feedback -- without needing to switch tools. This reduces the need of determining which specialized agent to use for each query, subscription cost, and system-switching friction.

\paragraph{Deploying Organizations} 
These are business entities that integrate agents into their products or operational pipelines to generate revenue, reduce costs, and enhance user engagement. These organizations consume agent capabilities at scale to generate business value, whether by embedding agents in their customer-facing products or using them to streamline internal operations. For example, healthcare providers may use diagnostic agents to improve initial assessment efficiency, while financial institutions might deploy advisory agents to offer personalized investment guidance to more customers while reducing personnel expenses associated with human financial advisors. Generalizable agents allow organizations to support multiple domains or shifting strategic focuses with minimal retraining~\cite{liu2025generalist, bhaskaran2025enterpriseai}. For instance, a financial institution could use one agent to manage both customer support and compliance monitoring~\cite{vukovic2025ai}.

\paragraph{Platform Owners} 
These are organizations that establish and maintain the operational spaces where LLM-based agents function, such as digital platforms, websites, or physical spaces equipped with agent interfaces. For example, travel aggregator platforms offer environments with ready-made databases and transaction systems that booking agents can leverage. Generalizable agents continue functioning despite environment changes such as UI redesigns~\cite{zheng2024gpt}, reducing maintenance costs and update coordination overhead~\cite{guo2024transagent}. Agent generalizability enables platform owners to maintain their platforms with minimal agent-specific customizations and less communication cost with agent developers; consequently, platform owners can develop platform accommodating multiple agents from different developers simultaneously, attracting more deploying organizations without incurring the prohibitive costs of creating and maintaining separate platform/environment variants or communication costs for different non-generalizable agents.


\subsubsection{Model Providers and Data Providers}
\phantom{for format correctness use}
\paragraph{Model Providers}
These stakeholders build and distribute backbone LLMs through APIs, checkpoints, or platforms~\cite{touvron2023llama, jiang2024mixtral}. A key future direction is building ``agent-friendly'' LLMs that integrate smoothly with diverse configurations of other agent components, i.e. the tool unit, the memory unit, and the perception and processing unit~\cite{anthropic2024mcp, shi2025tool}. Such compatibility supports consistent performance across heterogeneous agent deployments and reduces the need for repeated customization~\cite{shi2024ehragent, TiDB_2024}.

\paragraph{Data Providers} 
These entities collect, curate, and supply datasets used for training the backbone LLMs. Current datasets only optimize the backbone LLM's planning capabilities but overlook other agent components and their interactions. More specifically, most datasets designed for agent-related tasks only include pairs of user instructions and suggested final agent actions~\cite{moshkovich2025beyond, shridhar2020alfworld}, which solely optimize the backbone LLM's action planning capability while treating other components as immutable black boxes, assuming fixed configurations of agent specialized components that are determined before deployment rather than learned through data~\cite{shi2025tool, shi2025prompt}. This limits optimal information flow among agent components, such as how much webpage information should be preserved versus summarized based on specific user tasks, how many similar examples should be retrieved, and how tool specifications should be filtered and formatted based on task relevance~\cite{shi2025tool}. A promising direction is to create ``agent-friendly'' datasets that encode optimal component configurations -- including observation processing, memory retrieval, and tool use -- supporting end-to-end training that simultaneously tunes all agent components to optimize information flow between components~\cite{agentUmf, shi2025tool}. For example, a medical diagnosis dataset would include structured entries with user requests (``evaluate patient with chest pain and fever''), tool selection (diagnostic APIs), observations (which EHR data elements to perceive and process, and how to represent medical imagery), related memory knowledge (relevant historical cases), and step-by-step action traces showing component interactions. Each training data point would capture optimal configurations for all agent components -- such as the perception and processing unit prioritizing specific vital signs, the tool unit activating relevant diagnostic APIs, and the memory unit retrieving similar cases.

Training all agent components jointly from such datasets would allow agents to dynamically adjust configurations and internal strategies based on task requirements rather than relying on pre-configured settings that may be suboptimal for specific scenarios. For instance, the memory unit could learn to apply stricter similarity thresholds in regulatory domains and looser ones in creative domains, enhancing generalizability and security.

Data and model providers benefit directly from agent generalizability: successful generalizable agents drive recurring usage of backbone LLMs and datasets, increasing revenue and competitiveness.

\subsubsection{Agent Developers}
\label{sec:agent_developer}

Agent developers integrate backbone LLMs and agent specialized components from different providers into cohesive systems~\cite{liu2023agentbench, jain2022hugging}. The primary role is to integrate these components by establishing communication protocols, such as standardized data exchange formats or message passing systems~\cite{anthropic2024mcp}, between these heterogeneous components and ensuring their effective coordination within the agent architecture~\cite{agentUmf}.

As stated earlier, the lack of communication channels and orchestration of heterogeneous agent components needs to be addressed to achieve seamless integration of all agent components. As agent developers, one potential solution is to post-train the backbone LLM to dynamically configure other components based on task requirements. For example, the backbone LLM could be fine-tuned on instruction-tool pairs that teach it to effectively utilize tools given user instructions~\cite{qin2023toolllm}. However, such approaches may still be problematic, as potential mismatches can exist between the tools suggested by the backbone LLM and the actual available tools stored in the tool unit due to agent component heterogeneity. A straightforward method for agent developers is to post-train the backbone LLM using only the available tools in the specific tool unit, enabling better integration between the backbone LLM and the tool unit through improved understanding of tool usage patterns and tool selection strategies.

An interesting future direction for agent developers is to shift from isolated post-training of the backbone LLM to holistic training of the entire LLM-based agent, including agent specialized components and their interactions. This integrated approach could be implemented through two potential methodologies. 

Firstly, agent developers could implement a central coordination unit that optimizes component configurations independently of the backbone LLM~\cite{agentUmf}. Concretely, the central coordination unit -- implemented as a lightweight LLM, a neural network, or a rule-based system -- receives a task description and outputs an optimal component configuration (i.e., tool-selection policy, perception granularity, memory-retrieval parameters). It can be trained on task-configuration pairs, enabling the agent to adapt its tool use, observation processing, and memory strategies to the demands of each.

The second direction is the adoption of multi-agent architectures, where each specialized component functions as an autonomous agent or specialized LLM (e.g., an LLM specialized for tool use~\cite{qin2023toolllm} or memory retrieval~\cite{wang2023augmenting})~\cite{wang2024mixture, guo2024large}. In such systems, developers implement communication protocols that allow these components to negotiate, exchange information, and adapt to one another's representations~\cite{liu2024dynamic} to resolve potential incompatibilities caused by heterogeneity, enabling more flexible and scalable integration.

Agent generalizability maximizes developer return on investment (ROI) by reducing maintenance costs and expanding market reach. A generalizable agent can adapt to new domains with minimal fine-tuning, avoiding expensive data curation and LLM usage charges during retraining.

\subsubsection{Regulators and Policy-Makers}

These stakeholders are responsible for establishing the legal, ethical, and operational boundaries that govern the development and deployment of LLM-based agents~\cite{alanoca2025comparing}. Their oversight ensures that agent technologies align with societal values, public safety, and existing legal frameworks.

One key concern for regulators is generalizability. Agents must demonstrate consistent performance across demographic groups, geographic regions, and task contexts. For instance, the U.S. Food and Drug Administration (FDA) may require that diagnostic agents maintain comparable accuracy across diverse patient populations and clinical environments~\cite{windecker2025generalizability}. Without such consistency, agents risk introducing bias or unreliability in high-stakes domains like healthcare or criminal justice.

Another critical concern is security and privacy. Regulators aim to prevent agents from leaking sensitive data or enabling malicious behaviors. For example, restrictions may be placed on agents' access to vulnerability databases to avoid generating exploitation code~\cite{sadhu2024athena}, or on training data that includes copyrighted material without proper licensing~\cite{kanza2024geospatial}. These constraints directly influence how agents are trained, what information they can process, and how they are deployed in real-world systems.

Looking ahead, regulators may require agents to pass standardized generalizability benchmarks prior to deployment, particularly in domains that impact public welfare. In parallel, transparency requirements -- such as thorough model documentation and built-in interpretability mechanisms -- are likely to become essential for regulatory compliance. These tools will help ensure that agent behavior is not only generalizable but also explainable and auditable.

Future regulatory frameworks may formalize these expectations, mandating generalizable, and, more broadly, trustworthy agent behavior to align with broader social and legal norms.

%% file: sections_updated/measure_updated.tex
\section{Measuring LLM-based Agent Generalizability}\label{sec:measure_generalizability}

Measurement of generalizability of LLM-based agents is important and requires standardized datasets, evaluation protocols, and metrics. In this section, we review representative datasets (Section~\ref{sec:dataset}), evaluation (Section~\ref{sec:dimension2test_generalizability}), and metrics (Section~\ref{sec:generalizability_metric}) for measuring agent generalizability, followed by current limitations and future directions (Section~\ref{sec:measurement_limitation}).

\subsection{Datasets}\label{sec:dataset}
Datasets form the foundation for measuring the generalizability of LLM-based agents, as they define the contexts, tasks, and interaction against which performance is evaluated. A diverse set of benchmarks ensure that evaluations capture the breadth of environments in which agents may operate. In this subsection, we survey representative datasets across navigation, household, web, and operating system domains, highlighting their relevance for assessing generalizability.

\subsubsection{Navigation}
Navigation datasets test agents' ability to interpret and act on spatial instructions. They often evaluate generalizability to linguistic ambiguity, environmental diversity, and realistic interaction challenges.

\paragraph{Disfluent Navigational Instruction Audio Dataset (DNIA)~\cite{sun2024beyond}} DNIA is a navigation dataset that evaluates agents' ability to handle ambiguous spoken commands. It features both indoor and outdoor navigation scenarios with spoken instructions containing natural disfluencies such as ambiguous word choice, speech repair, and hesitation signs, specifically designed to test navigation agents' robustness to realistic communication challenges.
    
\paragraph{RoboTHOR Simulation Environment~\cite{deitke2020robothor}} RoboTHOR is an indoor navigation dataset that provides diverse household settings for visual navigation tasks. It features realistic indoor environments including kitchens, bedrooms, bathrooms, and living rooms with furniture (TV stands, dining tables) and 32 object types (mugs, laptops, TVs, apples), where agents start from random locations and navigate towards specified target objects.
    
\subsubsection{Household}
Household data benchmark agents in performing everyday domestic tasks, such as embodied 3D interactions. They capture both functional complexity and robustness to adversarial perturbations.  

\paragraph{VirtualHome~\cite{puig2018virtualhome}} VirtualHome is a household activity simulation dataset that models complex domestic tasks through atomic action sequences. It contains 2,821 programs covering 75 atomic actions and 308 objects, generating 2,709 unique steps for complex household activities (e.g., ``make coffee,'' ``watch TV,'' ``fold laundry'') with sequences of atomic actions (walk, grab, open/close, sit/standup) applied to household objects including furniture and appliances.
    
\paragraph{ALFWorld~\cite{shridhar2020alfworld}} ALFWorld is a dataset that combines text-based and visual 3D household interactions. It integrates interactive TextWorld~\cite{cote2018textworld} environments with embodied worlds from the ALFRED~\cite{shridhar2020alfred} dataset, providing both text-based and visual 3D household environments. Agents receive natural language instructions (e.g., ``put a washed apple in the kitchen fridge'') and must output either text commands (``go north,'' ``take apple'') in TextWorld or visual actions in 3D environments, completing household activities like cooking, cleaning, and object manipulation across kitchens, bedrooms, and living rooms in both text and visual modalities.

\paragraph{EIRAD~\cite{liu2024exploring}} EIRAD is a dataset containing 1,000 adversarial instances across scenarios involving navigation (such as room-to-room movement, obstacle avoidance, or waypoint following) and manipulation (such as object grasping, placement, or tool use) tasks. Adversarial instances include untargeted attacks that are designed to disrupt agent behavior like ``navigate to the kitchen with adversarial prompt suffix causing random erratic behavior,'' while targeted attacks manipulate agents toward specific unintended actions such as ``pick up the red cup with adversarial prompt suffix redirecting to `break the window instead'\,'' or ``move to the living room with prompt suffix forcing agent to `steal the jewelry'\,'' that redirect agents to perform predetermined malicious behaviors. 
    
\subsubsection{Web}
Web-based datasets focus on tasks such as online shopping, information retrieval, and website navigation. They test agents' ability to handle complex, multi-step interactions in diverse, dynamic digital environments.  

\paragraph{WebShop~\cite{yao2022webshop}} WebShop is a simulated e-commerce environment that measures agent performance on realistic online shopping tasks. It contains 1.18 million real-world products and 12,087 text instructions (with 1,600 human demonstrations of ground truth trajectories) where agents need to navigate multiple webpages to find, customize, and purchase items. Instructions are composed as complex queries like ``I'm looking for a quick-release replacement fitness strap band; it should match my chic teal fitbit, and price lower than 40.00 dollars,'' requiring agents to understand specifications, search effectively, filter results, and complete transactions.
    
\paragraph{Mind2Web~\cite{deng2024mind2web}} Mind2Web is a web-based dataset that measures agents' ability to follow instructions for complex web tasks. It contains 2,350 tasks (such as ``find one-way flights from New York to Toronto'') collected from 137 websites spanning 31 web-based domains (e.g., travel, info, service, shopping, entertainment), designed to test agents' capability to complete complex tasks on any website following natural language instructions.
    
\paragraph{WebArena~\cite{zhou2023webarena}} WebArena is a web environment designed for building and measuring autonomous web agents. It includes websites from four common domains: e-commerce, social forum discussions, collaborative software development, and content management. It includes diverse, long-horizon web-based tasks (with human annotated ground truth answers) that focus on measuring functional correctness of task completions, designed to emulate tasks that humans routinely perform on the internet, such as ``set up a new, empty repository with the name `\texttt{awesome\_llm\_reading}'\,''.
    
\subsubsection{Operating System}
Operating system benchmarks evaluate agents in realistic multi-application computing environments. They capture challenges in cross-platform generalization, system-level reasoning, and task execution.  

\paragraph{OSWorld~\cite{xie2024osworld}} OSWorld is a comprehensive computer environment that simulates multiple operating systems for agent evaluation. It provides Ubuntu, Windows, and macOS operating systems with essential applications including Chrome, VS Code, LibreOffice suite, GIMP, and basic OS applications like terminal and file manager. Tasks include complex cross-platform activities such as document editing in LibreOffice and email management in Thunderbird, designed to evaluate agents' ability to interact with diverse operating system environments.

\subsection{Establishing Evaluation Dimensions for Agent Generalizability}\label{sec:dimension2test_generalizability}
Before evaluating the specific \textit{dimensions} of agent generalizability, it is essential to establish the scope/boundary of an LLM-based agent. Here, a dimension refers to a particular axis along which an agent's ability to generalize can be systematically assessed, such as variation in user instructions, tasks, environments, or domains. To make such evaluations meaningful, a standardized and widely accepted framework is needed to define the scope of generalizability. One approach is a hierarchical domain-task ontology, in which broad domains are decomposed into sub-domains in increasingly specialized task categories. Construction of this ontology could leverage established industry classification system. For instance, the North American Industry Classification System (NAICS)~\cite{naics2022} provides a hierarchical framework categorizing economic activities from broad sectors (Information, Education, Healthcare) down to specific industries (Web Hosting within Computing Infrastructure). Existing benchmarks like Mind2Web~\cite{deng2024mind2web} can then provide task-level classifications within these industries, organizing web-based tasks by user intents, objects, and conditions. Within this structure, an agent's generalizability can be formally defined as the region in which it consistently demonstrates high task performance, thereby providing a clear boundary for evaluation.

Without such a framework, evaluations risk being inconsistent and difficult to compare. Current studies often claim generalizability at different levels of granularity, leading to ambiguity. For example, both Zhang et al.~\cite{zhang2024lamma} and Sun et al.~\cite{sun2024trustnavgpt} evaluate household agents, yet their definitions of generalizability differ substantially. The former focuses on object manipulation tasks (e.g., pick-up, place, open, close) across varied indoor environments, while the latter considers both indoor household navigation and outdoor navigation. Although both works claim household generalizability, their evaluation contexts are fundamentally different. Applying the same benchmark, such as a dataset emphasizing object manipulation, would unfairly favor one agent over the other, resulting in biased and non-comparable outcomes.

A standardized framework would resolve this issue by requiring claims of generalizability to be explicitly situated within a structured domain-task hierarchy. For instance, one study could be classified as \textit{household domain $\rightarrow$ object manipulation $\rightarrow$ indoor}, while another could be classified as \textit{household domain $\rightarrow$ household navigation}. With such a taxonomy, the scope of each agent becomes clearly defined, appropriate evaluation methods can be selected, and meaningful comparisons can be made. Moreover, it becomes straightforward to assess whether one agent is more generalizable than another by examining whether its defined region in the framework encompasses that of its counterpart.

Building on this foundation, we propose evaluating agent generalizability across four dimensions of increasing difficulty. First, \textbf{instruction-level generalizability} measures robustness to linguistic variation when the task and environment are fixed. For example, an agent should handle both the formal query ``book a flight to San Francisco'' and the colloquial request ``I gotta get my butt to the Bay Area,'' which ultimately refer to the same task. Second, \textbf{task-level generalizability} assesses the agent's performance on unseen tasks within the same environment, such as a web agent trained extensively on e-commerce tasks (like product purchasing) being tested on information retrieval tasks from search engines or knowledge-based websites, which require different interaction patterns despite operating in the same web environment. Third, \textbf{environment-level generalizability} evaluates consistency on the same task across structurally different environments. For instance, an agent trained in WebArena~\cite{zhou2023webarena} should be able to perform the same set of tasks consistently in Mind2Web~\cite{deng2024mind2web}. Both environments are web-based but differ significantly: WebArena uses self-hosted, simulated websites, while Mind2Web operates on actual live websites. Similarly, an agent trained on American Airlines' website should generalize to structurally distinct airline booking platforms such as Delta or Southwest. Finally, \textbf{domain-level generalizability} captures the ability to transfer core skills across domains/sub-domains. For example, an e-commerce agent adept at product purchasing and returns should also be capable of booking hotels and processing refunds in the travel domain, as these tasks share transferable competencies such as form submission, website navigation, and adherence to user preferences.

\subsection{Evaluation Metrics}\label{sec:generalizability_metric}
To evaluate the generalizability of LLM-based agents, it is first necessary to consider the metrics traditionally used to assess task-specific correctness/accuracy. These foundational measures provide the basis for more comprehensive evaluations of agent generalizability across varied settings.

\textbf{Success Rate (SR)}~\cite{liu2023agentbench}, also referred to as Task Goal Completion Rate (TGC)~\cite{trivedi2024appworld} or Valid Response Rate (VRR)~\cite{xie2024can}, quantifies the proportion of tasks in which an agent meets all evaluation criteria. Validation procedures depend on tasks and are often rule-based~\cite{zhu2024reliable}. For example, in a flight booking scenario, success requires satisfying constraints such as budget and destination, while in recommendation tasks, success may be measured by Recall@k, i.e., whether suggested items rank among the top-$k$ ground truth items. More advanced frameworks employ LLM-based judges or human annotators to assess outcomes~\cite{bai2024mt}. A representative example is Anthropic's HH-RLHF dataset, in which annotators conduct pairwise comparisons to select preferred responses based on helpfulness and harmlessness~\cite{bai2022training}.

Moving beyond binary success measures, \textbf{Step Success Rate (Step SR)} evaluates multi-step tasks by measuring the proportion of intermediate steps executed correctly, typically assessed by human evaluators. For instance, in the Mind2Web benchmark~\cite{deng2024mind2web}, Step SR includes both accurate identification of webpage elements (e.g., locating search fields) and correct action execution (e.g., typing input rather than clicking erroneously). In this example, achieving step-level success requires correctness in both component selection and action implementation.

In addition, \textbf{Human Alignment Scores} provide preference-based evaluation rather than binary judgments. Here, human annotators rate agent outputs against structured criteria using rating scales. For example, in dialogue evaluations, researchers sample multi-turn interactions and recruit experts to assess overall response quality on a 1-10 scale~\cite{bai2022training}. These scores capture nuanced aspects of quality, such as fluency, coherence, and user satisfaction.

\paragraph{Overall Score.}
To directly measure agent generalizability, \textbf{Overall Score} is the most widely adopted aggregate metric~\cite{yang2024agentoccam, zeng2023agenttuning}. It measures balanced performance across diverse task categories by computing a weighted average of category-specific success rates. For example, in web-based benchmarks, Overall Score may average results across categories such as online shopping, administrative tasks, social media interactions, and navigation tasks. Weighting schemes vary: some assign weights proportionally to the number of samples per category~\cite{yang2024agentoccam}, while others use customized weights to reflect task importance or difficulty~\cite{zeng2023agenttuning}. For instance, multi-step transactional tasks may receive higher weights than simple information retrieval tasks.

Higher Overall Scores indicate stronger agent generalizability, as they reflect the agent's ability to maintain consistent performance across heterogeneous task types rather than excelling narrowly in specific areas. The underlying scores for each task category are typically derived from task-specific metrics such as success rate or step success rate, depending on the nature of the task. For example, ALFWorld~\cite{shridhar2020alfworld} uses Success Rate to measure household task completion, WebShop~\cite{yao2022webshop} employs Human Alignment Scores to evaluate how well recommended products meet user requirements, and Mind2Web~\cite{deng2024mind2web} applies Step Success Rate to capture correctness across multi-step web navigation sequences. Together, these task-level measures and their aggregation into Overall Score provide a practical, though imperfect, basis for measuring the generalizability of LLM-based agents.

\subsection{Limitations and Future Directions in LLM-based Agent Generalizability Evaluation}\label{sec:measurement_limitation}

While existing metrics provide valuable tools for measuring agent performance, they also exhibit significant limitations when applied to generalizability. Current practices often conflate narrow task-specific accuracy with broad generalizability, overlook disparities across task categories, and fail to capture the tradeoffs between specialization and generalization. To advance the field, it is essential to identify these shortcomings and explore new directions that enable fairer, more comprehensive, and scientifically grounded evaluation. In the following subsections, we highlight three major challenges and outline potential solutions.

\subsubsection{Standardized Framework for Defining Agent Generalizability Boundaries}
A central obstacle in evaluating generalizability is the absence of a standardized, universally accepted framework to define its boundaries. Such a framework would enable testing methods for agent generalizability (for instance, the use of datasets) to cater to each agent's defined boundaries. Without such a framework, studies often adopt inconsistent definitions and levels of granularity, making comparisons across agents ambiguous and unfair. For example, consider two agents with overlapping but distinct expertise: one focuses on object manipulation in both indoor and outdoor environments, while another addresses indoor navigation and indoor object manipulation. A common dataset may allow comparison on the overlapping indoor object manipulation tasks, but without an explicit framework, claims of generalizability remain vague and difficult to interpret. 

A standardized framework would address this problem by explicitly situating each agent's capabilities within a structured taxonomy, thereby ensuring evaluations are aligned with the scope of the agent. For instance, one agent could be classified as \textit{household domain $\rightarrow$ object manipulation $\rightarrow$ indoor}, while another as \textit{household domain $\rightarrow$ navigation $\rightarrow$ indoor}. Such a taxonomy would make it possible to identify overlaps, and determine whether one agent is more generalizable than another by checking whether its defined region encompasses that of another agent. 

As a future direction, we propose that this framework adopt a hierarchical domain-task ontology, where broad domains are decomposed into sub-domains and progressively specialized task categories. Within this structure, an agent's generalizability can be formally defined as the region in which it consistently demonstrates high performance. This would enable meaningful, transparent, and reproducible evaluations and comparisons across agents.

\subsubsection{Variance-Based Metrics for Performance Disparities}
As discussed, current evaluation practices primarily rely on aggregate scores like Overall Score, computed as weighted averages of success rates across task categories. While convenient, these aggregate scores can mask substantial disparities in performance. An agent may achieve a seemingly competitive Overall Score (e.g., 43.1\%) by performing well primarily on tasks with higher weights, yet perform very unevenly across task categories -- for instance, scoring 61.3\% on social forum opinion exchange tasks but only 14.6\% on content management tasks~\cite{yang2024agentoccam}. Because content management tasks represent only 5.91\% of the training data, their impact on the Overall Score is minimal, even though the poor performance indicates a lack of true generalizability~\cite{mao2024robust}. 

The focus on learning high-weighted tasks can also inadvertently harm performance on lower-weighted tasks. Consider a conversational agent trained on a dataset where sentiment analysis tasks receive high weights (likely due to abundant labeled data) while mathematical reasoning tasks receive low weights (due to scarce training examples). The agent becomes highly proficient at sentiment analysis, but when faced with mathematical problems, it may inappropriately apply its sentiment analysis reasoning to mathematical tasks, leading to erroneous outputs~\cite{gupta2025llm}. 

To address this limitation, we propose complementing overall score with a simple yet fundamental variance-based measure. Specifically, the standard deviation of performance across task categories can expose disparities:
\begin{equation}
\sigma = \sqrt{\frac{1}{n}\sum_{i=1}^{n}(A_i - \bar{A})^2}
\end{equation}
where $n$ is the number of task categories, $A_i$ is the agent's accuracy on category $i$, and $\bar{A}$ is the weighted overall accuracy (Overall Score). A lower $\sigma$ indicates more consistent performance, while a higher $\sigma$ highlights disparities. 

A robustly generalizable agent should achieve both high Overall Score and low performance variance, regardless of their assigned weights. One limitation of this metric is that it cannot distinguish between uniformly poor agents (e.g., 10\% across all tasks) and uniformly strong ones (e.g., 85\% across all tasks). Thus, variance-based evaluation must be used alongside Overall Score to ensure both consistency and competence are captured.

\subsubsection{Evaluation Metrics for Generalizability Cost}
A further challenge lies in quantifying the tradeoff between task-specific accuracy and cross-task generalizability. As an illustration, consider a ticket-booking agent: a specialized agent may achieve 95\% accuracy on airline reservations but perform poorly on hotels or car rentals. In contrast, a more generalizable agent designed to handle multiple booking domains (flights, hotels, restaurants, events) may achieve only 80\% accuracy on airline bookings but maintain 75\% accuracy across all domains. This tradeoff is pervasive: as agents broaden their scope, performance on specific tasks often declines due to resource sharing across domains~\cite{zeng2023agenttuning}.

Current evaluation practices rarely capture this tradeoff explicitly, as they typically assess either narrow task accuracy or multi-task generalizability in isolation. This gap limits principled decision-making when choosing between specialization and generalization. Without such metrics, researchers and practitioners lacks principled ways to determine optimal generalizability levels for specific applications (i.e., designing agents to achieve generalizability with minimal sacrifice in task-specific accuracy), to evaluate whether generalization gains justify performance sacrifices on specific tasks, and to compare agents based on their efficiency in balancing specialization and generalization. 

To address this, we propose the \textit{Generalizability Cost (GC)} metric, defined as the average performance sacrifice when extending an agent's capabilities from a narrow specialist to a broader generalist:
\begin{equation}
GC = \frac{1}{n} \sum_{i=1}^{n} \big(A_i^{specialist} - A_i^{generalist}\big)
\end{equation}
where $A_i^{specialist}$ is the accuracy of a task-specific specialist on task $i$, and $A_i^{generalist}$ is the accuracy of a generalist agent on the same task. An intuitive way to think of this is that the task-specific specialist is an agent trained with a dataset solely on the current task, while the generalist agent is trained with a dataset spanning all $n$ tasks.

A lower $GC$ indicates that the generalist retains strong task-specific performance while achieving generalizability, whereas a higher $GC$ signals greater sacrifices in specialization. This metric would enable quantifying the efficiency of generalization, making informed decisions about the desired level of generalizability based on an acceptable task-specific performance sacrifice, and to facilitate comparisons between agents with different generalizability-accuracy tradeoffs.

%% file: sections_updated/improve_generalizability_llm.tex
\section{Improving LLM-Based Agent Generalizability via the Backbone LLM}
\label{sec:improve_generalizability_LLM}

The backbone LLM serves as the reasoning core of an LLM-based agent and is responsible for determining the actions the agent should take~\cite{huang2024understanding}. These actions are typically conveyed as function calls with associated arguments and executed by the tool unit~\cite{wang2024tools}. Consequently, the generalizability of the backbone LLM -- defined as its ability to consistently generate accurate plans across user instructions, tasks, environments, and domains~\cite{wang2024adapting, zeng2023agenttuning, deng2024mind2web} -- is a critical factor in determining the overall generalizability of the agent.

Approaches to improving the generalizability of the backbone LLM fall into two broad categories based on the LLM's lifecycle: a) strategies applied during the \textit{training stage}, and b) methods employed during the \textit{inference stage}. During training, efforts typically focus on diversifying instruction-tuning datasets across domains such as web navigation, household management, operating systems, and databases~\cite{zeng2023agenttuning, deng2024mind2web}, or similar domains containing different tasks, such as training a web-based agent using data from diverse tasks (online shopping, travel ticket booking, online health services, etc.) collected across a wide range of websites~\cite{deng2024mind2web}. During inference, recent work focuses on refining planning algorithms to generate robust action sequences across both seen and unseen environments~\cite{yao2022react, shinn2024reflexion, zheng2024gpt, zhang2024agent, sadhu2024athena, xiang2024guardagent, hu2024hiagent}.

\subsection{Backbone LLM's Training Stage}

Efforts to enhance the backbone LLM during training can be divided into two major dimensions: a) the design of the training data~\cite{zeng2023agenttuning, deng2024mind2web} and b) the optimization of the training objectives~\cite{zeng2023agenttuning, deng2024mind2web, wang2024adapting}.

\subsubsection{Training Data Design}

The generalizability of the backbone LLM is closely tied to the diversity and relevance of its training data. Rather than training on a narrow set of tasks, the data should span multiple domains and task types to help the model generalize across different interaction modalities and planning contexts.

Several techniques for general-purpose LLM training apply. For example, \textit{curriculum learning} ranks data from simple to complex to gradually improve generalization~\cite{soviany2022curriculum}, while \textit{data selection} strategies help identify the most valuable data subsets~\cite{albalak2024survey}. \textit{Data mixing}~\cite{raffel2020exploring, gao2020pile} combines datasets from multiple domains with optimized sampling weights.

To improve the generalizability of agent-specific LLMs, recent work has curated diverse instruction-tuning datasets across multiple domains. \citet{deng2024mind2web}, for instance, propose a dataset with over 2,000 open-ended tasks drawn from 137 websites across 31 domains. Their work emphasizes the use of real-world rather than simulated data, capturing a wide spectrum of user interaction patterns to expand the scope of possible tasks, such as enabling users to engage in sophisticated ways with websites, as opposed to basic operations such as searching, following links, and reading content. Building on this, \citet{zeng2023agenttuning} introduce a multi-task dataset spanning six agent benchmarks -- ALFWorld~\cite{shridhar2020alfworld}, WebShop~\cite{yao2022webshop}, Mind2Web~\cite{deng2024mind2web}, Knowledge Graph, Operating System, and Database~\cite{liu2023agentbench}. They also use GPT-4 to generate additional instructions and procedural steps, filtered using task success rewards to ensure high-quality samples.

In addition to domain diversity, training data may include multiple modalities. \citet{wang2024adapting} highlight that both linguistic feedback (e.g., textual comments) and non-linguistic reward signals (e.g., scalar success indicators) can be effective components of instruction-tuning data. While earlier studies~\cite{li2023camel, micheli2021language} rely primarily on textual feedback and use reward signals only for data filtering, \citet{wang2024adapting} propose incorporating both modalities as training data -- learning from linguistic feedback via unsupervised methods and maximizing rewards from non-linguistic inputs.

\subsubsection{Training Objective Design}

A complementary approach to improving backbone LLM generalizability focuses on refining the training objective. Specifically, the loss function should not be overly influenced by a small set of tasks or domains but should instead reflect the diverse operational needs of the agent~\cite{deng2024mind2web}.

To this, \citet{zeng2023agenttuning} introduce a dual-part training set structure: a) an agent-specific dataset covering tasks such as household operations and web navigation, and b) a general-purpose LLM dataset that includes math reasoning, coding, and multi-domain natural language understanding. They design a mixed-objective loss function that jointly optimizes performance on both datasets and observe that training exclusively on agent-specific data degrades agent generalizability. In contrast, the mixed objective enhances performance across both agent-specific and general tasks.

This insight also extends to modality balancing. As discussed, \citet{wang2024adapting} propose that objective functions account for both linguistic and reward-based feedback during instruction-tuning, rather than optimizing for one modality in isolation. This multi-signal objective encourages the backbone LLM to extract complementary cues and improves its ability to generalize across modalities and task formats.

\subsubsection{Limitations and Future Directions}

Despite recent progress, several challenges remain in designing generalizable training protocols for backbone LLMs. One key limitation is the reliance on predetermined, static training datasets. Current fine-tuning frameworks often do not dynamically adapt based on the model’s evolving performance. As a result, the model may encounter similar challenges without exposure to examples targeting its weaknesses~\cite{deng2024mind2web, zeng2023agenttuning, kiela2021dynabench}.

To address this, \citet{qi2024webrl} propose a dynamic task generation framework that identifies agent failure cases and generates new tasks with slightly altered parameters to target those weaknesses. While promising, their method is designed for online, interactive environments that can introduce latency due to real-time feedback loops, as the system must wait for responses from actual environments, making the training process prohibitively time-consuming. A promising future direction involves adapting such dynamic curricula to sandbox environments~\cite{ruan2023identifying}, which enable faster iterations without real-world execution overhead, while still capturing the benefits of adaptive training.

\subsection{Backbone LLM's Inference Stage}

During the inference stage, the backbone LLM must reliably generate accurate and context-appropriate plans across diverse and potentially unseen tasks and environments~\cite{wang2024adapting, zeng2023agenttuning, deng2024mind2web}. To this end, existing methods~\cite{yao2022react, zhang2024lamma} primarily focus on improving the agent's planning algorithms to enhance generalizability during real-time decision-making.

\subsubsection{In-Context Learning with Prompted Examples}

A commonly adopted strategy for improving generalizability during inference is leveraging the in-context learning capabilities of the backbone LLM~\cite{yao2022react}. This approach involves providing the LLM with task-relevant examples -- either embedded directly into the user instruction as \textit{direct prompts}~\cite{zheng2024gpt, zhang2024agent}, or retrieved from the memory unit as \textit{indirect prompts}~\cite{sadhu2024athena, xiang2024guardagent, hu2024hiagent}. These examples help the model infer task structure and expected action formats, even when confronted with previously unseen scenarios.

\subsubsection{Structured Planning with Domain-Invariant Representations}

Beyond in-context learning, a promising direction involves using structured planning frameworks that explicitly separate \textit{domain-invariant planning logic} from \textit{environment-specific details}. The core idea is to decompose the planning process into two parts:

\begin{itemize}
    \item \textbf{Universal action schemas and planning rules}, which remain constant across environments, tasks and domains. For example, in household tasks, the \texttt{pick-up(object)} action always requires the object to be graspable and the agent to be co-located, and the \texttt{move(from, to)} action requires a traversable path.
    \item \textbf{Environment-specific facts}, which vary for each new environment (e.g., which objects exist in the current environment, their initial spatial relationships, and the specific goal configuration).
\end{itemize}

This separation allows the backbone LLM to reuse learned planning logic across environments.

Consider a ``retrieve object'' task in two different environments: a household kitchen and a space station. Without structured planning, the backbone LLM might generate environment-specific action plans such as:

\begin{itemize}
    \item Kitchen: ``walk to the counter, open the drawer, grab the spatula, close the drawer.''
    \item Space station: ``execute EVA protocol, depressurize airlock, engage magnetic boots, navigate to external storage pod, activate retrieval arm sequence.''
\end{itemize}

While these sequences differ greatly in environment-specific facts, they share the same action schemas and planning rules (such as the \texttt{pick-up} action always requiring the object to be accessible and the agent to be co-located, regardless of environment) and a standardized planning format (such as \texttt{(:objects ...) (:init (inside ...)) (:goal (holding ...))}). Moreover, the planning logic remains consistent: identifying the target object, determining its location, and applying predefined actions (\texttt{pickup}, \texttt{move}) according to established preconditions. By representing both environments using a standardized planning language, such as the Planning Domain Definition Language (PDDL)~\cite{jiang2019task}, the backbone LLM can maintain planning consistency. For instance:

\begin{itemize}
    \item Kitchen PDDL: \texttt{(:objects spatula1 drawer1) (:init (inside spatula1 drawer1)) (:goal (holding spatula1))}
    \item Space Station PDDL: \texttt{(:objects tool1 storage-pod1) (:init (inside tool1 storage-pod1)) (:goal (holding tool1))}
\end{itemize}

In both cases, the universal planning rule and planning logic -- what \texttt{pick-up} means and when it applies -- remains identical across both scenarios; only the specific objects and their locations change. This consistency enables the agent to transfer the same action schemas, rules, planning format, and planning logic to different environments without requiring domain-specific adaptation or retraining, thereby improving agent generalizability.

\subsubsection{Case Study: Generalization via PDDL in Household Environments}

\citet{zhang2024lamma} exemplify the structured planning approach in household environments. The method uses PDDL to define universal action schemas -- such as \texttt{pick-up}, \texttt{open-door}, and \texttt{move} -- in a fixed domain file that is shared across all environments. For each specific environment (e.g., a new household floor plan), the backbone LLM generates a corresponding PDDL problem file describing the environment-specific objects, their initial locations, and the desired goal configuration.

A PDDL planner then performs the combinatorial search over these representations to generate a valid and optimized action sequence. This method improves agent generalizability by enabling consistent performance across a wide variety of unseen household layouts and object arrangements. The key strength lies in decoupling static planning logic from dynamic world states and using a standardized symbolic representation to structure the agent's reasoning process.

%% file: sections_updated/improve_generalizability_specialized_component.tex
\section{Improving LLM-based Agent Generalizability via Agent Specialized Components}
\label{sec:improve_generalizability_agent_components}

The generalizability of LLM-based agents can be improved through more effective design of the agent's specialized components, including the perception and processing unit, the memory unit, and the tool unit. While generalizability ultimately depends on coordinated interaction among all components, improving the capabilities of each individual component can significantly enhance the agent's ability to operate across varied tasks, environments, and domains.

\subsection{The Perception and Processing Unit}

The perception and processing unit plays a critical role in capturing and transforming environmental input for the backbone LLM. Generalizability in this component can be improved by enhancing two key capabilities: \textit{observation perception} and \textit{observation processing}~\cite{yang2024agentoccam, yang2024towards}.

\subsubsection{Observation Perception}

Observation perception refers to the agent’s ability to detect and extract meaningful signals from its environment~\cite{yang2024agentoccam}. For instance, in autonomous driving, LiDAR-based object detection enables agents to perceive road obstacles and dynamic objects~\cite{song2024enhancing}. Depending on the task, agents need to perceive information across different modalities and settings. For instance, a web-based agent~\cite{deng2024mind2web, yang2024agentoccam} needs to operate across diverse websites (e.g., booking platforms, e-commerce portals, social media pages), while a household agent~\cite{sun2024trustnavgpt, zheng2024evaluating, zhang2024lamma} must navigate environments with varied spatial layouts and object configurations.

Most current efforts in observation perception focus on improving generalizability across similar environments (e.g., generalizing across websites)~\cite{yang2024agentoccam, yang2024towards}. However, generalizing across fundamentally different domains remains an open challenge. Future work should explore how agents can identify the boundaries of their operational domains, rather than attempting to create universal observation perception systems that unrealistically stretch across disparate domains (such as expecting a single agent to generalize from cooking to driving). Designing perception systems that can recognize when environmental input is out-of-distribution is critical for ensuring task success and agent robustness.

Moreover, even for the same agent task, the associated environment may not remain static~\cite{yang2024towards}. Instead, the underlying dynamics of the environment can be complex and challenging to identify~\cite{tamkin2022task}. For instance, an invalid video link that was previously working may cause the agent to fail in retrieving necessary content, requiring it to adapt its strategy by searching for alternative sources or notifying the user about the issue. These dynamics may also be influenced by temporality~\cite{zhou2024hazard}, which refers to changes in the environment over time, such as the periodic display of announcements, or stochasticity~\cite{wu2023smartplay}, which describes transitions between environments interlaced with independent random events. For example, a shopping agent might encounter unexpected browser popups. An agent with high generalizability must consistently perceive relevant elements under such dynamics, adapting its perception strategy when conditions change or input degrades.

\subsubsection{Observation Processing}

Observation processing refers to transforming raw perceived input into structured formats that the backbone LLM can consume for planning. This includes tasks such as formatting HTML tables into Markdown, summarizing content, or compressing redundant structure tokens~\cite{yang2024agentoccam}. Unlike collecting raw environmental data, processing ensures that the input passed to the backbone LLM is concise, semantically rich, and task-relevant.

Effective observation processing improves agent generalizability by filtering out irrelevant or misleading content (e.g., distracting banners or unrelated links) that may confuse the subsequent action planning process and preserving essential features necessary for planning. For example, \citet{yang2024agentoccam} propose simplifying observations while maintaining key information, allowing agents to generalize more effectively across varied websites and tasks. This is especially important for memory-based planning, where stored observations indirectly serve as in-context samples~\cite{sumers2023cognitive, kotseruba202040}.

\subsection{The Memory Unit}

The memory unit contributes to agent generalizability through \emph{management flexibility} and \emph{data storage and utilization}.

\subsubsection{Management Flexibility}

Memory management flexibility refers to the ability to easily insert, update, and delete stored content~\cite{xiang2024guardagent}. This adaptability allows the agent to maintain an up-to-date internal representation of task-relevant knowledge, ensuring that retrieved memories reflect current task requirements and environmental constraints. This flexibility directly enhances generalizability by enabling rapid adaptation to unseen tasks. For example, consider an agent trained primarily on airline ticket booking that encounters a new task: hotel reservation. When faced with unfamiliar hotel booking interfaces and terminology, the agent allow flexible insertion of relevant examples into its memory unit -- such as sample hotel search queries, booking confirmation patterns, or successful reservation workflows from similar hospitality domains. The backbone LLM can then retrieve these newly inserted samples during planning, using them as contextual references as guidance on the newly hotel booking task. 


\subsubsection{Data Storage and Utilization}

This dimension focuses on what information the agent stores and how it uses that information to guide future decisions. Memory units typically store both short-term memory (e.g., user instructions, agent thoughts, observations, feedback) and long-term memory (e.g., task-relevant documents or procedural guides in external knowledge bases)~\cite{sumers2023cognitive, kotseruba202040}. These memories can be leveraged to support in-context learning, where the backbone LLM generalizes to unseen tasks by grounding its reasoning in past examples~\cite{zhang2024agent}.

Regarding knowledge stored in the memory unit,
\citet{wang2024adapting} emphasize the importance of diversity -- storing both linguistic feedback and non-linguistic reward signals, which helps the agent adapt to varied task types. However, not all stored data contributes positively to planning. \citet{yang2024agentoccam} find that irrelevant memory can reduce generalizability, highlighting a better interaction between the memory unit and the backbone LLM (discussed in Section~\ref{sec:selective_data_utilization_in_memory_unit}).

\subsection{The Tool Unit}

The tool unit, which consists of callable external functions, enhances agent generalizability by increasing the agent's capacity to interact with the environment and solve a broader range of problems. We discuss two important design factors, \emph{tool diversity} and \emph{management flexibility}.

\subsubsection{Tool Diversity}

Tool diversity refers to equipping the agent with a wide array of function calls that enable it to handle a variety of operations, rather than just incorporating tools for specific tasks~\cite{xiang2024guardagent}. For example, in web-based environments, agents should not be limited to basic actions like reading content or following links. Instead, they should support more complex interactions such as clicking, selecting, typing, and form submission~\cite{deng2024mind2web}. Incorporating such diverse capabilities allows agents to generalize across different web domains and application types.

\subsubsection{Management Flexibility}

Similar to the memory unit, the tool unit benefits from dynamic management capabilities. Users should be able to insert new tools, update existing ones, or remove outdated functions. When faced with new tasks that require operations beyond the current toolset, management flexibility enables quick adaptation -- for instance, by uploading code for new tools -- thus enhancing the agent’s generalizability to previously unseen tasks~\cite{xiang2024guardagent}.

%% file: sections_updated/improve_generalizability_interaction.tex
\section{Generalizability of Interactions between the Backbone LLM and Agent Specialized Components}\label{sec:improve_generalizability_interaction}

Recall that after an agent executes an action, the resulting observation is perceived and processed by the agent~\cite{song2024enhancing, yang2024agentoccam}. These perceived observations are then stored in the memory unit to support future task execution~\cite{sumers2023cognitive, kotseruba202040}. When encountering a new task, relevant knowledge retrieved -- including observations from similar past tasks -- is transferred to the backbone LLM to guide its planning and decision-making. For example, in a new ticket booking scenario on Airline $B$'s website, the agent may retrieve prior experience from Airline $A$, such as observations including a ``button `Apply Rewards'\,'' or ``text `Member Discounts'\,''. The transfer of such information -- perceived observations, prior actions, and task-specific knowledge -- constitutes the interaction between the backbone LLM and specialized components.

Recent studies~\cite{yang2024agentoccam, hu2024hiagent} have focused on enhancing generalizability by managing these inter-component interactions. Specifically, they examine the transition of data from agent specialized components (e.g., the memory unit) to the backbone LLM. This includes the flow of current observations and stored knowledge -- such as previous thoughts, actions, and results relevant to the ongoing task -- into the backbone LLM to support adaptive planning.

Managing this transition like controlling which data should be transferred and which should not is critical because transferring all potentially relevant information without filtering can negatively affect generalizability. As noted by \citet{yang2024agentoccam}, indiscriminately passing every observed or stored detail to the backbone LLM increases planning complexity, inflates inference costs, and reduces the agent's ability to generalize to new tasks or environments. For instance, suppose an agent stores both a static text ``My Account'' and an identically worded clickable link in its memory. When the agent encounters a new website with an account access section, the memory unit transfers both the non-actionable, non-interactive static text and the actionable link as contextual guidance for the backbone LLM's action generation process. However, this creates a fundamental problem for the backbone LLM's planning capabilities: it must now determine which of these identical but functionally different elements to target when generating actions for account-related tasks like editing user profiles on unseen websites. When retrieved context contains both interactive and non-interactive elements but with the same content, the backbone LLM needs to disambiguate between them and correctly identify the actionable elements. However, when disambiguation fails -- choosing the static text over clicking the link as the current action -- the subsequent actions generated may become non-functional, leading the agent to incorrectly conclude that the task is impossible when the correct interactive elements (link) were available.

To mitigate such issues, it is essential to manage both the storage of observations in the memory unit (i.e., the transition from the perception and processing unit to the memory unit) and the retrieval of relevant knowledge from memory (i.e., the transition from the memory unit to the backbone LLM). Selectively transferring only functionally relevant elements -- such as the clickable link -- instead of irrelevant misleading noise ensures the backbone LLM receives unambiguous, actionable information. This improves its ability to generate accurate and effective action sequences across varying and previously unseen environments, thereby enhancing the overall generalizability of the agent.

In this section, we further analyze these interaction pathways by categorizing them into two key types: a) interaction between the perception and processing unit and the memory unit, and b) interaction between the memory unit and the backbone LLM. Each of these pathways presents unique challenges and opportunities for improving the transfer of task-relevant knowledge while minimizing environment-specific noise.


\subsection{Interaction between the Perception and Processing Unit and the Memory Unit}

When an agent performs an action, it perceives environmental observations through the perception and processing unit. These observations are then transferred to the memory unit, where they are stored as references for future, similar tasks~\cite{sumers2023cognitive, kotseruba202040}. During planning, the memory unit retrieves stored observations that share semantic or structural similarity with the current task. However, unfiltered storage of observations can significantly degrade generalizability~\cite{yang2024agentoccam, hu2024hiagent}, as the retrieved memory may include both transferable task-relevant information and environment-specific noise.

For example, if an agent stores all perceived elements (e.g., pop-up advertisements) during a ticket booking session, these irrelevant details may be retrieved in future booking tasks. When applied to a different airline's interface, such noise (e.g., promotional banners or loyalty program widgets) may not only confuse the backbone LLM but also mislead its action planning. The core issue lies in the following: when every detailed observation is captured and stored without filtering, the memory unit becomes populated with both generalizable task-relevant patterns (such as input fields for dates, selection buttons for options, and form submission elements) and environment-specific irrelevant details (such as promotional banners, site-specific branding, and decorative visual elements). When a new but similar task queries the memory unit for relevant historical observations, it retrieves this mixed content, forcing the backbone LLM to process and distinguish between transferable patterns and environmental noise during planning.

\begin{example}\label{ex:environmentnoise}
Consider an agent that first books a flight on Delta Airlines' website. The perception unit captures both essential booking components (e.g., ``Departure City,'' ``Search Flights'' button) and Delta-specific artifacts (e.g., ``SkyMiles Rewards,'' ``FLASH SALE: 50\% off to Hawaii''). When the agent later plans a booking on United Airlines' website, the memory unit retrieves the Delta-related content due to task similarity, which contains both relevant generalizable patterns and irrelevant Delta-specific environmental noise due to unfiltering. However, because the Delta-specific details are not relevant to United's interface, the backbone LLM may waste actions searching for nonexistent ``SkyMiles'' sections on United's website or attempt to apply Delta-specific interaction patterns to United's interface. This undermines generalizability by conflating transferable booking logic with irrelevant site-specific content.
\qed
\end{example}

Example~\ref{ex:environmentnoise} highlights a critical insight: not all perceived information should be stored in memory. Instead, managing the data flow from the perception and processing unit to the memory unit is essential to filter out task-irrelevant elements before they pollute the historical record and interfere with the agent's planning on new tasks. Reducing the amount of unnecessary information ensures that future retrievals present the backbone LLM with only task-relevant content, thereby minimizing inference complexity and improving cross-environment generalizability.

The core principle here is to reduce observation-level noise before it enters the memory system. When the agent indiscriminately stores every observed detail, subsequent tasks that rely on memory retrieval risk being overloaded with redundant or misleading information. This not only saturates the backbone LLM's context window but also strains its attention mechanisms, making it more difficult to focus on the essential cues needed for planning.

To address this, several methods have been proposed to filter observations based on task relevance. For instance, \citet{yang2024agentoccam} present a tree-structured parsing strategy, where observed content is organized into a hierarchical layout. The agent identifies ``pivotal nodes'' -- critical elements for tasks, such as a clickable ``View Order'' link to access order details -- and selectively stores these nodes along with their structural context. Specifically, the agent retains:
\begin{itemize}
    \item \textbf{Ancestor nodes}, indicating where the pivotal element resides (e.g., the parent table of order information);
    \item \textbf{Sibling nodes}, providing immediate context in the same table row (e.g., order number and total price);
    \item \textbf{Descendant nodes}, offering detailed characteristics like the clickable link within ``View Order.''
\end{itemize}
This structured filtering ensures that observations preserve functional relationships needed for task generalization while omitting decorative headers and ads that vary widely across environments but don't affect the core task logic.

Complementarily, \citet{ji2024dynamic} propose a relevance-based approach, where observations are selected based on their embedding similarity to the agent's current action and previously stored knowledge. Only those segments of the observation space that strongly correlate with the task goal are retained, ensuring that only useful content is preserved.

In summary, managing the interaction between the perception and processing unit and the memory unit is vital for maintaining the agent's generalizability. By ensuring that only task-relevant information is stored -- while filtering out environmental noise -- the agent can more reliably transfer learned patterns across diverse tasks and interfaces.

\subsection{Interaction between the Memory Unit and the Backbone LLM}
\label{sec:selective_data_utilization_in_memory_unit}

While the previous subsection focused on filtering task-irrelevant observations before they are stored in the memory unit, another critical aspect of improving agent generalizability involves managing what information is retrieved from the memory unit and passed to the backbone LLM during task planning~\cite{hu2024hiagent}. Specifically, instead of transferring all previously stored data related to similar tasks, the agent should selectively retrieve only the most relevant knowledge -- such as task-relevant thoughts, actions, and observations -- that supports the current planning objective. This filtering process ensures that the backbone LLM receives compact, meaningful signals and avoids being overwhelmed by environment-specific or contextually irrelevant information.

\begin{example}
Consider an agent performing a seat selection task on Delta Airlines' website. The memory unit may store rich, detailed records, including thoughts like ``need to select window seat in rows 12-15 for better comfort,'' actions such as \texttt{[click(``seat\_map''), locate(``window\_seats''), select(``row\_14A''), \\ click(``blue\_confirm\_seat'')]}, and observations like ``Delta seat map shows rows 12-15 available, blue `Confirm Seat' button in bottom right, \$25 upgrade fee for preferred seating.'' If this full memory is transferred to the backbone LLM during a new seat selection task on United Airlines' website, the backbone LLM may erroneously attempt to locate Delta-specific UI elements -- like a ``blue Confirm Seat'' button or a ``rows 12-15'' -- which do not exist in the new environment. This can lead to failed action sequences and task abandonment.

By contrast, if the memory unit instead transfers a compressed, abstracted summary -- e.g., ``Thought: Selected preferred seating based on comfort criteria. Action: Navigated seat map and chose window seat with upgrade fee consideration. Observation: Successfully selected desired seat type with associated cost'' -- the backbone LLM can apply the same planning logic to United's interface and execute analogous actions like \texttt{[click(``choose\_seats''), select(``window\_seat''), confirm(``selection'')]}. Even if the seat map layout or row numbering differs, the backbone LLM retains the transferable decision pattern (e.g., prioritizing comfort and handling fee-based upgrades), resulting in successful adaptation to a new environment. This illustrates how summarization and abstraction of past experiences enables the agent to preserve essential planning patterns while discarding environment-specific details that would otherwise prevent successful adaptation to unseen environments.
\qed
\end{example}

Two strategies have been proposed to manage this interaction between the memory unit and the backbone LLM. The first approach, introduced by \citet{yang2024agentoccam}, involves decomposing tasks into structured sub-tasks organized in a tree hierarchy. During planning, the backbone LLM receives only the data associated with sub-tasks that share a common parent with the current sub-task. The paper suggests that it is not the entire previous task stored in the memory unit that is relevant to the current sub-task; instead, it is only part of the previous task (some sub-tasks) that is relevant and should be retrieved. For instance, in flight booking, if the current sub-task is seat selection, the agent retrieves only prior seat selection-related experiences rather than irrelevant segments such as baggage selection or payment processing. This sub-task-aware filtering ensures that the retrieved data is aligned with the current planning context.

The second approach, proposed by \citet{hu2024hiagent}, takes a complementary perspective: instead of retrieving only parts of a previous task, the agent retrieves a compressed summary of the entire prior task. This summarization can be generated using LLMs or a dedicated module, which distills complex interactions into concise representations. For example, rather than providing the full sequence of actions and observations during a previous booking session, the agent may retrieve a high-level narrative like: ``Completed booking by selecting lowest-cost fare with upgrade options based on user preference, followed by selecting economy window seat and confirming purchase.'' Such high-level summaries retain planning-relevant information while abstracting away environment-specific implementation details.

Both approaches aim to prevent the backbone LLM from being overloaded with irrelevant data that may not generalize across contexts. By supplying the backbone LLM with summarized memories and transferable patterns, the agent is better positioned to adapt its planning strategies across new environments, thereby improving its generalizability.

%% file: sections_updated/generalizable_method.tex
\section{Generalizable Frameworks vs. Generalizable Agents: Differences and Connections}\label{sec:generalizability_method_vs_generalizable_agent}

Although many frameworks have been proposed to improve the performance of LLM-based agents, their relationship to agent-level generalizability remains unclear. In particular, it is important to distinguish between \textit{generalizable frameworks} -- frameworks that can consistently achieve strong performance when applied to a variety of scenarios (user instructions, tasks, environments, or domains) when combined with the associated fine-tuning data -- and \textit{generalizable agents} -- agents that exhibit consistently high performance across diverse, unseen scenarios. Clarifying this distinction is crucial, as deploying a framework that generalizes well does not necessarily guarantee that the resulting agent will generalize across different, especially unseen, scenarios. This section examines the differences between these two concepts, surveys representative generalizable frameworks, and discusses how methodological advances can be leveraged to construct agents with broader and more reliable generalization capabilities.

\subsection{Differences Between Generalizable Frameworks and Generalizable Agents}

With the rapid rise of LLM-based agents, a wide range of frameworks have been proposed to improve their performance. However, \textit{performance} is a multidimensional concept that spans accuracy on specific tasks, generalizability across tasks and domains, security against adversarial inputs~\cite{zhang2024agent, li2024personal}, protection against privacy leakage~\cite{zhan2024injecagent, xie2024recall}, fairness~\cite{zhang2024sa}, and more. A key gap in the literature is that existing studies rarely specify which aspect of performance a given proposed framework seeks to improve. In particular, the connection between individual frameworks and the broader goal of agent generalizability remains largely unexplored.

To address this, we introduce the notion of \textbf{generalizable frameworks}. A generalizable framework is one that produces consistently high performance regardless of instruction, task, environment, or domain type. For example, a framework that achieves strong accuracy on household tasks when combined with household fine-tuning data can also yield strong accuracy on web tasks when combined with web-based fine-tuning data. In other words, when combined with appropriate fine-tuning data for a specific task, a generalizable framework enables an agent to perform that task well. Notable examples include frameworks such as ReAct~\cite{yao2022react} and Reflexion~\cite{shinn2024reflexion}.

However, using a generalizable framework does not guarantee that the resulting \emph{agent} will itself be generalizable. We define a \textbf{generalizable agent}, as discussed earlier, as one that maintains consistently high performance across varied instructions, tasks, environments, or domains, especially those not in the agent's fine-tuning data. For instance, an agent fine-tuned solely on household object pickup tasks should be able to generalize to related actions such as placement, opening, and closing, and even to other sub-domains like cooking or cleaning, without additional fine-tuning.  

The distinction is therefore crucial:  
\begin{itemize}
    \item A \emph{generalizable framework} ensures strong performance on any specific scenario when fine-tuned on data for that scenario, but does not inherently transfer to scenarios outside the fine-tuning data distribution.  
    \item A \emph{generalizable agent} exhibits robustness across different and unseen scenario, achieving consistent accuracy beyond the scope of its fine-tuning data.  
\end{itemize}

This contrast highlights a central research challenge: how to bridge the gap between methodological generalizability and agent-level generalizability -- i.e., how to design frameworks that not only perform consistently, but also enable agents to generalize across diverse, dynamic environments.

To frame this challenge, we categorize potential generalizable frameworks that aim to improve agent performance (e.g., accuracy) into six themes:
\begin{itemize}
    \item User Instruction Optimization and Intent Capture (Section~\ref{sec:user_instruction_optimization}),  
    \item Iterative Action Generation (Section~\ref{sec:iterative_action_generation}),  
    \item Reflection (Section~\ref{sec:reflection}),  
    \item Uncertainty Analysis (Section~\ref{sec:uncertainty_analysis}),  
    \item Human Alignability (Section~\ref{sec:human_alignability}), and  
    \item Interactions Between the Backbone LLM and Agent Specialized Components (Section~\ref{sec:interaction_generalizable_method}).
\end{itemize}

These perspectives correspond to different stages of the agent workflow. Instruction optimization focuses on clarifying user requests before planning, while iterative action generation concerns the step-by-step decomposition of tasks. Reflection, uncertainty analysis, and human alignability serve as refinement mechanisms after planning, enabling agents to self-correct, quantify confidence, or align with human expectations. Finally, frameworks for improving interactions between the backbone LLM and agent specialized components ensure that planned actions can be faithfully executed. Together, these six categories provide a structured view of generalizable frameworks.

While we highlight promising generalizable frameworks, empirical benchmarks are still needed to evaluate their cross-scenario generalizability. Some frameworks may excel in one scenario (e.g., household tasks) but fail in another (e.g., web tasks).  Hence, a framework is generalizable if it consistently performs well across scenarios.


\subsection{User Instruction Optimization and User Intent Capture}\label{sec:user_instruction_optimization}
Accurate interpretation of user instructions is the first step in an agent's reasoning process. Before the backbone LLM can organize thoughts, set objectives, and determine actions, it must understand what the user is asking. The clarity and quality of user instructions are therefore strongly tied to agent performance. Ambiguous or incomplete instructions may lead to misinterpretation of user intent and suboptimal planning, whereas well-structured instructions -- containing clear task descriptions, relevant constraints, sample tool APIs, or reminders to incorporate memory and observations -- enable the agent to fully leverage its capabilities.

Optimizing user instructions has been shown to improve task performance. For instance, \citet{chen-etal-2024-prompt} propose an iterative method that ranks candidate instructions by quality, and refines them until optimal task-specific prompts are obtained. Beyond optimizing instructions themselves, capturing the \textit{user intent} is equally critical. Since intent is not always explicitly expressed in instructions~\cite{yang2024towards}, the agent's ability to infer it greatly impacts action performance. General LLM approaches typically achieve this by fine-tuning with human feedback~\cite{ouyang2022training}, rewarding responses that correctly identify and fulfill intent. Extending this idea to agents, \citet{yang2024towards} propose storing actions that successfully captured user intent in memory, allowing the agent to reuse them when encountering similar new instructions.

User instructions may also arrive in different modalities, such as text, audio, or images~\cite{sun2024trustnavgpt}. Current LLM-based agents~\cite{achiam2023gpt, ren2023robots} primarily process text, often overlooking intent cues in other modalities. For example, subtle features of human speech -- such as tone, pitch, or hesitation -- can signal ambiguity or uncertainty~\cite{prestopnik2000relations, golledge2003human}, but are typically discarded when audio is converted directly to text. This limitation can exacerbate hallucinations and reduce performance in non-text scenarios. To address this, \citet{sun2024trustnavgpt} develop an uncertainty modeling framework for audio-guided navigation agents, combining semantic uncertainty (e.g., ambiguous word choice, speech repair) with vocal uncertainty (e.g., pitch, loudness, speech rate). Future research should move beyond converting audio into text and instead allow LLMs to directly reason over raw multimodal signals, leveraging advances such as AudioLM~\cite{borsos2023audiolm} or Whisper~\cite{radford2023robust}. Similarly, incorporating visual cues, gestures, and contextual signals from images~\cite{ma2024caution} remains an open area for enhancing multimodal instruction understanding.

\subsubsection{Limitations and Future Directions}  
Several challenges remain in optimizing instructions and capturing intent.  

\paragraph{Leveraging Agent Specialized Components for Ambiguity Resolution} Most work relies solely on the backbone LLM to resolve ambiguities~\cite{chen-etal-2024-prompt, sun2024trustnavgpt}, overlooking cascading errors across components. For example, an ambiguous instruction such as ``Get me the report'' may trigger irrelevant tool use and misguided perception. Future research should explore how signals from perception, memory, and tool units can be combined to detect and resolve ambiguity. For instance, contextual observations from perception can help disambiguate vague references (e.g., clarifying ``remote control'' as ``television remote''), but more systematic cross-component strategies are needed beyond household scenarios.

\paragraph{Managing Cross-component Conflicts} Ambiguity may also produce conflicting interpretations across components. For example, a smart home agent instructed to ``Make it comfortable in here'' may receive contradictory signals: perception detects high room temperature (82°F), memory recalls a user preference for 72°F, and the user's clothing suggests they may already feel cold. Such conflicts highlight the need for explicit frameworks to classify ambiguity types and identify component-level contradictions. Future directions include human-in-the-loop clarification strategies that balance efficiency with user satisfaction, avoiding excessively burdensome clarification dialogues~\cite{zhang2023clarify, zhang2024modeling}.

In summary, user instruction optimization and intent capture are central to improving agent performance. Moving forward, research should focus on multimodal intent inference, ambiguity resolution through component integration, and conflict-aware clarification strategies to enable more robust and user-aligned instruction understanding.


\subsection{Iterative Action Generation}\label{sec:iterative_action_generation}

During planning, the backbone LLM typically decomposes a user's task into a sequence of sub-tasks~\cite{xi2025rise, huang2024understanding}. Many approaches adopt an \textit{iterative strategy}, where actions are generated step by step, with each subsequent action conditioned on prior reasoning and outcomes, allowing agents to adapt dynamically to observations and intermediate results.

A prominent example is the ReAct framework~\cite{yao2022react}, which integrates chain-of-thought (CoT) reasoning~\cite{wei2022chain} with environmental feedback. In ReAct, completing a multi-step task requires an alternating sequence of thoughts, actions, and observations. Building on this, \citet{aksitov2023rest} combine ReAct with Reinforced Self-Training~\cite{gulcehre2023reinforced} to create an iterative self-improving framework, where generated thought-action sequences are filtered and reused to fine-tune the backbone LLM, progressively enhancing planning capabilities.

Given this iterative nature, the quality of each planned action is critical. To improve action selection, several frameworks incorporate reward-based mechanisms. \citet{xiong2024watch} employ Monte Carlo sampling to estimate the reward of candidate actions, using deviations from correct actions as contrastive signals for training. Similarly, \citet{chen2024llm} leverage Monte Carlo Tree Search (MCTS)~\cite{browne2012survey}, where environment states are represented as nodes and actions as edges, enabling efficient exploration and evaluation of possible action sequences. Complementary to these approaches, \citet{liu2023reason} introduce Bayesian adaptive Markov decision processes~\cite{duff2002optimal} in which an auxiliary LLM models the reward of entire action sequences. The agent then selects the highest-reward sequence but executes only the first action, repeating this process iteratively until the task is completed.

\subsubsection{Limitations and Future Directions}  
Despite their effectiveness, existing iterative action generation methods face key limitations, particularly in \textit{long-horizon tasks} that involve extended action sequences with interdependencies and delayed rewards, such as web navigation~\cite{qi2024webrl} or device control~\cite{erdogan2024tinyagent}. Explicit reward modeling approaches~\cite{xiong2024watch, chen2024llm, liu2023reason} often suffer from high computational complexity and sparse reward signals~\cite{swiechowski2023monte}, which hinder responsiveness in real-time applications and may degrade plan quality.

Recent work has attempted to mitigate these issues by decomposing planning into high-level strategy generation followed by environment-specific execution~\cite{erdogan2025plan}, or by leveraging memory retrieval to support long-term reasoning~\cite{nguyen2024one}. However, these approaches generally lack explicit reward modeling for intermediate actions, limiting their ability to evaluate and adjust ongoing plans. A promising research direction is the development of \textit{dynamic reward-modeling techniques} that adapt to context. In some cases, rewards for intermediate actions may depend on the eventual outcome of the full sequence~\cite{liu2023reason}, requiring look-ahead reasoning. In others, rewards could be computed efficiently from intermediate results. Designing hybrid reward models that balance long-term outcome awareness with computational efficiency represents an important avenue for enabling accurate and scalable planning in long-horizon settings.

\subsection{Reflection}\label{sec:reflection}

Reflection refers to an LLM-based agent's ability to evaluate and revise its planned actions based on feedback~\cite{xi2025rise, huang2024understanding}. By incorporating feedback loops, agents can self-correct errors, refine action sequences, and improve task performance. Reflection mechanisms can be broadly categorized into two types: \textit{internal feedback} and \textit{external feedback}.

\textbf{Internal feedback} is generated within the agent, most commonly by the backbone LLM itself. This approach has been described as ``self-refine''~\cite{madaan2023self}, ``self-check''~\cite{miao2023selfcheck}, or ``self-reflection''~\cite{shinn2024reflexion}. Most frameworks focus on verifying the correctness of planned actions by prompting the LLM to critique or revise its own outputs~\cite{miao2023selfcheck, shinn2024reflexion}. Beyond correctness, \citet{wang2024executable} propose an internal reflection mechanism that verifies the \textit{format} of planned actions. For instance, agents may be required to output in a structured format such as ``User instruction: \ldots, Available APIs: \ldots.'' Such rigid formats, however, can restrict the action space, as the agent can only plan within the predefined API set. To address this limitation, their approach translates planned actions into executable Python code, which is then run by a Python interpreter -- expanding the effective action space while maintaining verifiability.

\textbf{External feedback} arises from the environment and provides signals about the success or failure of executed actions. This feedback may take the form of simple binary rewards~\cite{shinn2024reflexion} or more complex signals such as error messages from interpreters or APIs~\cite{huang2024queryagent}. To make better use of such signals, \citet{du2024anytool} introduce a hierarchical tool organization strategy. When an agent receives negative external feedback, it reflects by identifying which tool invocation failed within the hierarchy and re-plans using alternative tools. This structured approach enables agents to localize errors more effectively and recover from failures during execution.

In summary, reflection enhances the performance of LLM-based agents by enabling them to iteratively refine their plans in response to feedback. Internal reflection leverages the backbone LLM itself for self-critique or format verification, while external reflection relies on environmental signals to guide re-planning. Together, these mechanisms highlight the potential of feedback-driven refinement to improve performance in dynamic and uncertain environments.

\subsection{Uncertainty Analysis}\label{sec:uncertainty_analysis}

Uncertainty in the context of LLM-based agents refers to the degree of confidence an agent has in its interpretation of user instructions~\cite{sun2024trustnavgpt, ruan2023identifying}, its planning process~\cite{fang2024inferact}, or the outcomes of its actions~\cite{zheng2024evaluating}. Sources of uncertainty are diverse and include ambiguous instructions~\cite{sun2024trustnavgpt, ruan2023identifying}, noisy or incomplete observations~\cite{ma2024caution, song2024enhancing}, and inherent limitations in agent components, such as the backbone LLM's difficulty with long-horizon reasoning~\cite{erdogan2025plan, nguyen2024one} or the memory unit's tendency to retrieve irrelevant knowledge~\cite{yang2024agentoccam}.  

When properly quantified, uncertainty serves as a valuable signal for improving agent performance. It enables agents to decide when additional planning is needed~\cite{zheng2024evaluating} or when to request human intervention~\cite{fang2024inferact}, such as asking clarifying questions or seeking supplementary information. For example, when instructed to ``clean some soap bar and put it in the cabinet,'' an agent that mistakenly selects a soap bottle might detect low confidence (e.g., 10\%) in its action and prompt the user for clarification before execution~\cite{fang2024inferact}. This kind of preemptive uncertainty analysis prevents errors and allows agents to pause, seek guidance, or re-plan rather than executing potentially faulty actions.  

While uncertainty estimation has been extensively studied in general LLMs~\cite{geng2024survey, ren2023robots}, its application in LLM-based agents introduces additional complexity. Unlike general language models, agents can incorporate observations and environmental signals into uncertainty assessments. For instance, \citet{fang2024inferact} propose measuring uncertainty as the alignment between an inferred user task -- derived from observed action-observation sequences -- and the ground-truth task. Similarly, \citet{zheng2024evaluating} design an uncertainty-based failure detection module using multimodal LLMs (MLLMs) such as LLaVA~\cite{liu2024visual}, which take image-based observations as input and produce both a reliability assessment and an uncertainty estimate. Other approaches analyze uncertainty directly at the instruction level: \citet{sun2024trustnavgpt} quantify semantic ambiguity (e.g., vague wording, speech repairs) and vocal features (e.g., pitch, loudness, hesitation) in audio commands, embedding these signals into prompts for more cautious planning.

\subsubsection{Limitations and Future Directions}  
Current uncertainty analysis frameworks have important gaps:  

\paragraph{Limited Focus on Observations} Most work emphasizes uncertainty in instructions~\cite{sun2024trustnavgpt} or actions~\cite{fang2024inferact, zheng2024evaluating}, but overlooks uncertainty in observations. For example, a shopping agent asked to find the best audio system may be misled by distracting advertisements (noisy observations) or by perceiving products from only a single brand (incomplete observations). Existing studies~\cite{ma2024caution, song2024enhancing} show that corrupted observations can degrade performance or even lead to catastrophic failures in domains such as autonomous driving, highlighting the need for comprehensive observation-level uncertainty modeling.  

\paragraph{Interaction with Memory Retrieval} Planning typically involves retrieving prior actions and observations from the memory unit. The extent of this retrieval directly impacts uncertainty estimation: recalling only the most recent sub-task may yield insufficient context, while retrieving all historical information may overwhelm the model with irrelevant data~\cite{yang2024agentoccam}. Future work should investigate adaptive retrieval strategies that prioritize knowledge most effective at reducing uncertainty, expanding retrieval only when confidence thresholds are unmet.  

\paragraph{Scalability to Long-horizon Tasks} Many methods compute uncertainty over entire action-observation sequences~\cite{fang2024inferact, zheng2024evaluating}. While effective for short tasks, this becomes computationally expensive in long-horizon settings~\cite{liu2025uncertainty}. Localizing uncertainty to individual steps can reduce cost but risks missing global interdependencies. For example, an agent purchasing audio components might assign high confidence to each individual item, yet fail to detect incompatibilities among them (e.g., mismatched amplifier and speaker specifications). This tension suggests the need for hierarchical uncertainty frameworks that combine local assessments with selective global reasoning, analyzing only critical dependencies while avoiding full-sequence overhead.

In summary, uncertainty analysis enhances the performance of LLM-based agents by enabling proactive error detection and cautious decision-making. Future research should focus on integrating observation-level uncertainties, developing adaptive memory retrieval strategies, and designing scalable frameworks for long-horizon tasks.

\subsection{Human Alignability}\label{sec:human_alignability}

Human alignability refers to the extent to which an LLM-based agent can align its behavior with human reasoning, expectations, and values. This includes demonstrating human-like reasoning capabilities~\cite{xie2024can, su2024ai, tennant2024moral} and mirroring human learning by acquiring social norms and values through self-evolving interactions~\cite{liu2023training}. In contrast to systems that rely exclusively on pre-defined training signals, a human-alignable agent can autonomously improve its actions by inferring rewards, generating feedback beyond scalar ratings, and adapting based on environmental observations. Studies show that enhancing human alignability can improve not only interpretability but also general performance such as task accuracy~\cite{liu2023training, tennant2024moral, yang2024towards}.

\subsubsection{Alignability in General LLMs}  
In general LLMs, human alignability is commonly pursued through \textit{Supervised Fine-Tuning (SFT)} and \textit{Reinforcement Learning from Human Feedback (RLHF)}. SFT fine-tunes models on curated instruction-output pairs, offering computational efficiency~\cite{zhang2023instruction}, but suffers from high data-creation costs and limited generalizability to scenarios beyond the fine-tuning data~\cite{gudibande2023false}. RLHF instead trains a reward model from human feedback on LLM responses~\cite{bai2022training}, enabling large-scale alignment without manual grading of every response. However, RLHF faces challenges such as biases in human evaluators~\cite{santurkar2023whose, cotra2021ai} and difficulties in representing complex, diverse human values~\cite{stiennon2020learning}. For a detailed discussion of SFT and RLHF, readers are referred to recent surveys~\cite{casper2023open, zhang2023instruction}.

\subsubsection{Alignability in LLM-based Agents}  
LLM-based agents extend beyond general LLMs by integrating perception, memory, and tool-use capabilities, which open opportunities for more dynamic forms of alignment. \citet{liu2023training} propose a self-evolving approach where the backbone LLM not only generates human-aligned actions but also produces feedback on misaligned ones, effectively acting as a ``self-critic.'' During inference, the agent adjusts its behavior in real time using this feedback. Complementarily, \citet{tennant2024moral} design an agent that infers rewards directly from environmental payoffs (e.g., outcomes in an Iterated Prisoner's Dilemma), using these signals to fine-tune its policies toward human moral values.  

\subsubsection{Distinctive Features and Future Directions}  
Compared to general LLMs, which typically require explicit human intervention (e.g., evaluators providing feedback in RLHF~\cite{ouyang2022training}), LLM-based agents emphasize \textit{self-evolving alignment}. They can generate feedback automatically -- either via self-critique~\cite{liu2023training} or by inferring it from environmental signals~\cite{tennant2024moral}. This distinction is crucial for real-world deployment: agents often operate in unpredictable settings where pre-specified human feedback cannot fit every scenario, and reliance on human evaluators would introduce latency in time-sensitive applications~\cite{wang2024understanding}. Moreover, agent architectures naturally facilitate automated feedback generation, since observations from perception or knowledge retrieved from memory can serve as alignment signals.  

Looking forward, promising directions include developing methods that combine multiple feedback sources (e.g., perception, memory, and external interactions), exploring richer social learning mechanisms, and designing evaluation benchmarks that measure alignment not only to task accuracy but also to human values across diverse scenarios. By advancing these directions, LLM-based agents can move closer to robust, human-centered generalizability.

\subsection{Interactions between the Backbone LLM and Agent Components}\label{sec:interaction_generalizable_method}

Effective interactions between the backbone LLM and agent specialized components depends on the extent to which the backbone LLM is informed about their capabilities, limitations, and potential sources of error. Without this awareness, the backbone LLM may generate infeasible or unsafe plans, limiting the overall performance of the agent.  

After planning, the backbone LLM typically outputs actions as function calls with input arguments, which are then executed by the tool unit. However, as \citet{xiang2024guardagent} observe, the backbone LLM may attempt to invoke non-existent functions that are unsupported by the tool unit. This highlights the need for the backbone LLM to be explicitly aware of the scope of available tools.  

Further challenges arise because planned actions are often required to follow a predefined format~\cite{wang2024executable}, such as ``User instruction: \ldots, Available APIs: \ldots.'' While this improves consistency, it also constrains the agent's action space. Simply informing the backbone LLM of the existence of certain tools, without clarifying their specific functionalities, may prevent the system from composing tools effectively to solve complex tasks. As a result, the backbone LLM may overlook opportunities for multi-tool reasoning and generate overly narrow or suboptimal action plans~\cite{wang2024executable}.  

Another challenge stems from tool execution errors. Tools may fail, produce inconsistent outputs, or require specific conditions to function correctly~\cite{zhang2024breaking}. Robust planning therefore requires not only knowledge of tool scope and capabilities, but also of their potential failure modes. Consider a financial advisory agent asked to ``analyze my portfolio and suggest tax-optimized rebalancing.'' Without detailed tool awareness, the backbone LLM might simply invoke a generic \texttt{portfolio\_analysis()} function, unaware that it requires premium API access, or fail to recognize that combining \texttt{tax\_impact()} with \texttt{investment\_analysis()} would yield more accurate recommendations. Furthermore, if the \texttt{tax\_impact()} tool occasionally fails with large datasets, the backbone LLM would be unable to anticipate this or implement fallback strategies. With comprehensive tool awareness, however, the backbone LLM could design a more robust plan: first checking access privileges, then combining complementary functions, and finally incorporating error-handling routines.  

In summary, effective interaction between the backbone LLM and agent components requires more than enumerating available tools -- it requires informing the backbone LLM about tool functionalities, dependencies, and error profiles. Such awareness enables the backbone LLM to compose tools effectively, anticipate failures, and generate robust plans that enhance performance in complex applications.

\subsection{From Generalizable Frameworks to Generalizable Agents: Future Directions}

The field has produced many promising generalizable frameworks, often without explicitly framing them as such. However, the broader question of how methodological generalizability translates into agent-level generalizability remains largely unanswered. Bridging this gap requires systematic evaluation and principled integration of frameworks with agent-level design strategies. We highlight three key future directions.

\subsubsection{Benchmarks for Framework-level Generalizability} The first step is to establish benchmarks that evaluate whether a framework is truly generalizable across different scenarios (i.e., user instructions, tasks, environments, or domains). Some frameworks may work well in a single scenario (e.g., household tasks) but fail to transfer even when retrained on another (e.g., web-based tasks). To test generalizability, multiple training and testing datasets across diverse dimensions should be prepared. A framework can then be deemed generalizable if, when combined with scenario-specific fine-tuning data to train the agent, the agent consistently exhibits high performance on that scenario's test set. Such a benchmark would allow fair comparison across frameworks and clarify the extent to which each contributes to generalizability.

\subsubsection{Benchmarks for Agent-level Generalizability} A second line of work should assess whether agents equipped with generalizable frameworks themselves achieve generalizable behavior. This requires controlled experimental designs. For example, an agent using ReAct~\cite{yao2022react} or Reflexion~\cite{shinn2024reflexion} could be fine-tuned and evaluated on ``held-in'' benchmarks such as ALFWorld (household object manipulation) and BabyAI (grid-world navigation). Its performance would then be tested on ``held-out'' datasets such as HotPotQA (multi-hop reasoning) or GSM8K (mathematical problem solving). Consistently high performance across held-out datasets would provide empirical evidence that method-level generalizability can indeed yield agent-level generalizability.

\subsubsection{Synergistic Integration of Frameworks and Agent-level Techniques} Even if a framework demonstrates generalizability, it may not directly produce a generalizable agent. A promising direction is to combine generalizable frameworks with techniques specifically designed to enhance agent generalizability, such as diverse training data, component-level improvements, or better LLM-component interactions (see Sections~\ref{sec:improve_generalizability_LLM}, \ref{sec:improve_generalizability_agent_components}, and \ref{sec:improve_generalizability_interaction}). For instance, combining reflection (Section~\ref{sec:reflection}) with training diversity strategies has been shown to outperform either approach alone~\cite{fu2025agentrefine}, highlighting the potential of methodological synergy.

Taken together, these directions suggest the need for a third type of benchmark: one that evaluates combinations of generalizable frameworks and agent-level generalizability enhancement techniques. Such benchmarks would reveal whether hybrid approaches can achieve greater generalizability than any individual method, ultimately moving the field closer to agents that generalize reliably across diverse, real-world contexts.

%% file: sections_updated/conclusion_updated.tex
\section{Conclusion and Future Directions}\label{sec:conclusion_updated}

In this survey, we addressed a critical gap in the rapidly evolving field of LLM-based agents by providing the first comprehensive examination of \emph{generalizability} -- a fundamental property that determines whether these systems can consistently perform across diverse user instructions, tasks, environments, and domains. Through our systematic analysis, we established a formal definition of generalizability in LLM-based agents, examined why it matters from the perspectives of multiple stakeholders, introduced a structured taxonomy for measuring and improving generalizability, and clarified the distinction and connections between \emph{generalizable frameworks} and \emph{generalizable agents}.

Although we reviewed a rich body of literature, many important challenges remain open. Below, we highlight several key directions that we believe deserve further exploration.

\textbf{Novel Architectural Frameworks for Component Coordination.}  
We see promising opportunities in developing new agent architectures that improve coordination among backbone LLMs and agent specialized components. Current systems typically rely on the backbone LLM to directly orchestrate heterogeneous modules, which can limit generalizability. We propose investigating central coordination units that manage interactions across these components of different configurations, resolve incompatibilities, and optimize information flow. Future research should focus on designing, training, and integrating such frameworks to systematically enhance agent generalizability.

\textbf{Standardized Frameworks for Defining Agent Generalizability.}  
We also recognize the need for a standardized framework to define the boundaries of an agent's capabilities. At present, studies claim generalizability at vastly different levels of granularity. We propose investigating a hierarchical domain-task ontology to explicitly situate claims of generalizability, thereby enabling rigorous, transparent, and reproducible evaluation across agents.

\textbf{Benchmarks for Framework-to-Agent Generalizability Translation.}  
Finally, we emphasize the importance of benchmarks that assess both a) whether a framework is truly generalizable across various scenarios and b) whether deploying such a framework results in a generalizable agent. Moreover, we see value in benchmarks that assess combinations of generalizable frameworks with techniques for improving agent-level generalizability, allowing us to measure their joint effectiveness compared to using each approach in isolation.

As LLM-based agents transition from research prototypes to deployment in high-stakes domains such as healthcare, finance, and autonomous systems, we believe that addressing these challenges is essential for building systems that are generalizale, and more broadly, trustworthy. We hope that our analysis and proposed directions will serve as a foundation for future research, highlighting that systematic evaluation, architectural improvement, and methodological innovation are all necessary to realize the full potential of generalizable LLM-based agents in real-world applications.